\documentclass[lettersize,journal]{IEEEtran}

\usepackage{array}
\usepackage[caption=false,font=normalsize,labelfont=sf,textfont=sf]{subfig}
\usepackage{textcomp}
\usepackage{stfloats}
\usepackage{url}
\usepackage{verbatim}
\usepackage{graphicx}
\usepackage[numbers,sort&compress]{natbib}
\usepackage{amsmath,amssymb,enumerate,amsfonts}

\usepackage{amsthm}
\usepackage{mathtools}
\usepackage{breqn}
\usepackage{algorithm, algorithmicx, algpseudocode}

\usepackage{blindtext}
\usepackage{gensymb}
\usepackage{xparse}
\usepackage{lipsum}
\usepackage{mathrsfs}
\usepackage[mathscr]{euscript}
\usepackage{times}
\usepackage{balance}
\usepackage{multirow}
\usepackage{booktabs}
\usepackage[usenames,dvipsnames,svgnames,table, xcdraw]{xcolor}

\usepackage[colorlinks=true,pdfpagemode=UseNone,citecolor=black,linkcolor=black,urlcolor=BrickRed]{hyperref}
\usepackage{comment}

\newtheorem{theorem}{Theorem}

\newtheorem{remark}{Remark}

\newtheorem{definition}{Definition}

\newcommand{\squeezeup}{\vspace{-3mm}}

\newcommand{\transpose}{\mathsf{T}}
\newcommand{\m}{\mathop{\mathrm{m}}}

\newcommand{\Hz}{\mathop{\mathrm{Hz}}}

\usepackage[]{mdframed}

\begin{document}

\title{Proprioceptive Invariant Robot State Estimation}
% : Theory and Practice}

\author{{Tzu-Yuan Lin, Tingjun Li, Wenzhe Tong, Maani Ghaffari}
        % <-this % stops a space
\thanks{Funding for M. Ghaffari was in part provided by NSF Award No. 2118818. This work was also partly supported by the Amazon Research Awards.~\textit{(Corresponding author: Tzu-Yuan Lin.)}}
\thanks{The authors are with the University of Michigan, Ann Arbor, MI 48109, USA.~\texttt{\{tzuyuan, tingjunl, wenzhet, maanigj\}@umich.edu}.}
}



\maketitle

\begin{abstract}
% In this paper, we present DRIFT: Dead Reckoning for Robotics In Field Time, an open-source symmetry-preserving robot state estimation library that works in real-time on a variety of robotic platforms. The current implementation is based on the state-of-the-art invariant Kalman filtering, which we provide a didactic introduction. To enhance performance on low-cost robots, the framework comes with two optional modules, a contact estimator and a gyro filter. We perform  Furthermore, simulation results with a marine robot are reported. We evaluate the performance in many aspects, including one long-horizon experiment aiming to understand the limit of DRIFT. Throughout all experiments, DRIFT outperforms the baseline method and demonstrates the potential of being a robust odometry system.

This paper reports on developing a real-time invariant proprioceptive robot state estimation framework called DRIFT. A didactic introduction to invariant Kalman filtering is provided to make this cutting-edge symmetry-preserving approach accessible to a broader range of robotics applications. Furthermore, this work dives into the development of a proprioceptive state estimation framework for dead reckoning that only consumes data from an onboard inertial measurement unit and kinematics of the robot, with two optional modules, a contact estimator and a gyro filter for low-cost robots, enabling a significant capability on a variety of robotics platforms to track the robot's state over long trajectories in the absence of perceptual data. Extensive real-world experiments using a legged robot, an indoor wheeled robot, a field robot, and a full-size vehicle, as well as simulation results with a marine robot, are provided to understand the limits of DRIFT. 

\end{abstract}


\section{Introduction}

Autonomous robots can greatly benefit humanity by taking over dangerous and tedious jobs, such as extraplanetary exploration, rescue missions in disaster scenes, warehouse logistics, and daily home assistance for elders~\cite{Carroll2023Nasa}. In order to accomplish the above tasks, autonomous robots must be able to navigate reliably in various environments and maintain stability. Modern motion planning and control algorithms rely heavily on accurate estimations of the robot's states, i.e., orientation, position, and velocity. Subsequently, accurate state estimation is a key capability in robot autonomy. 

\begin{figure}
\centering 
\subfloat{\includegraphics[width=0.99\columnwidth]{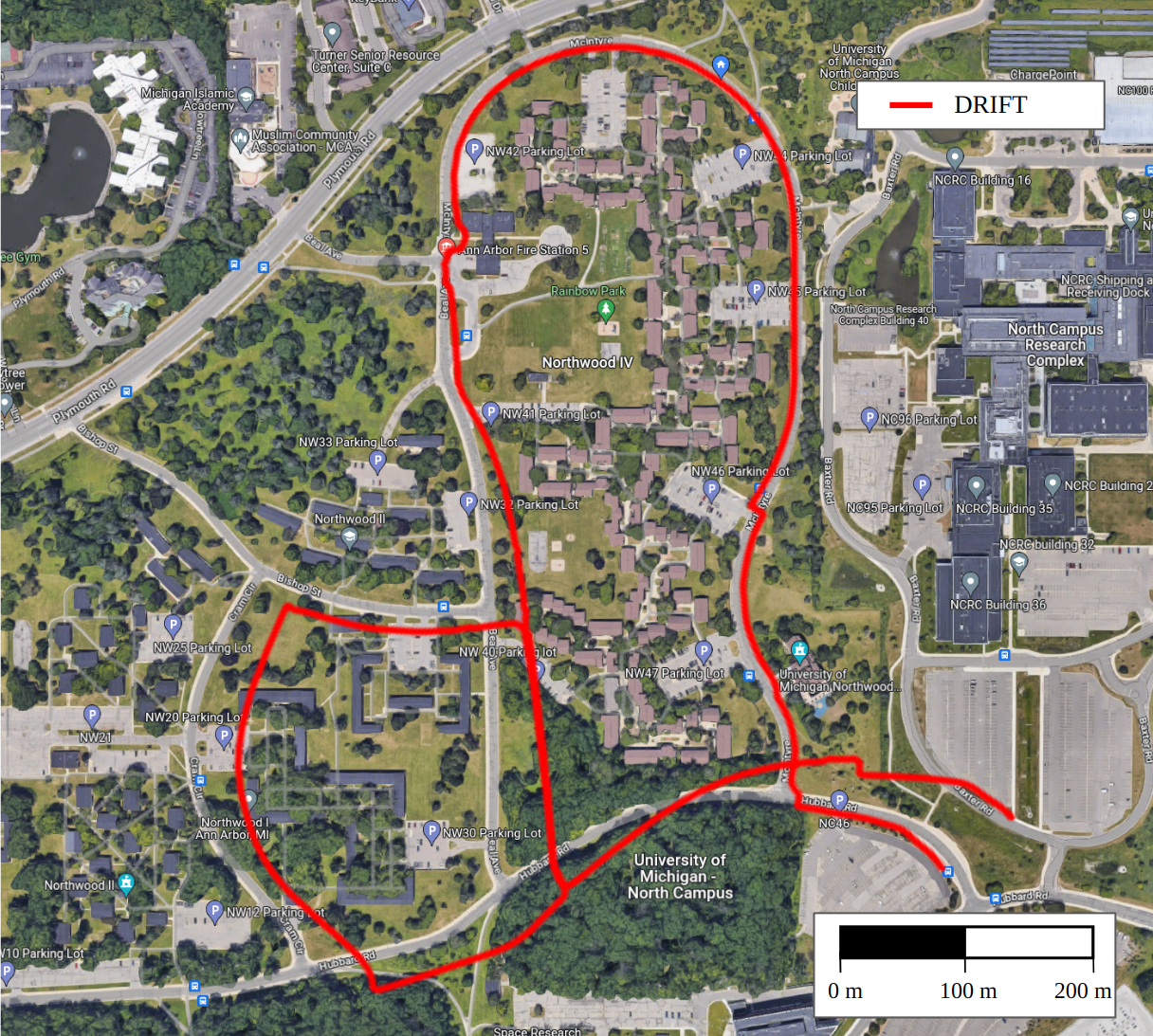}}\\
\vspace{-3.3mm}
\subfloat{\includegraphics[width=0.99\columnwidth]{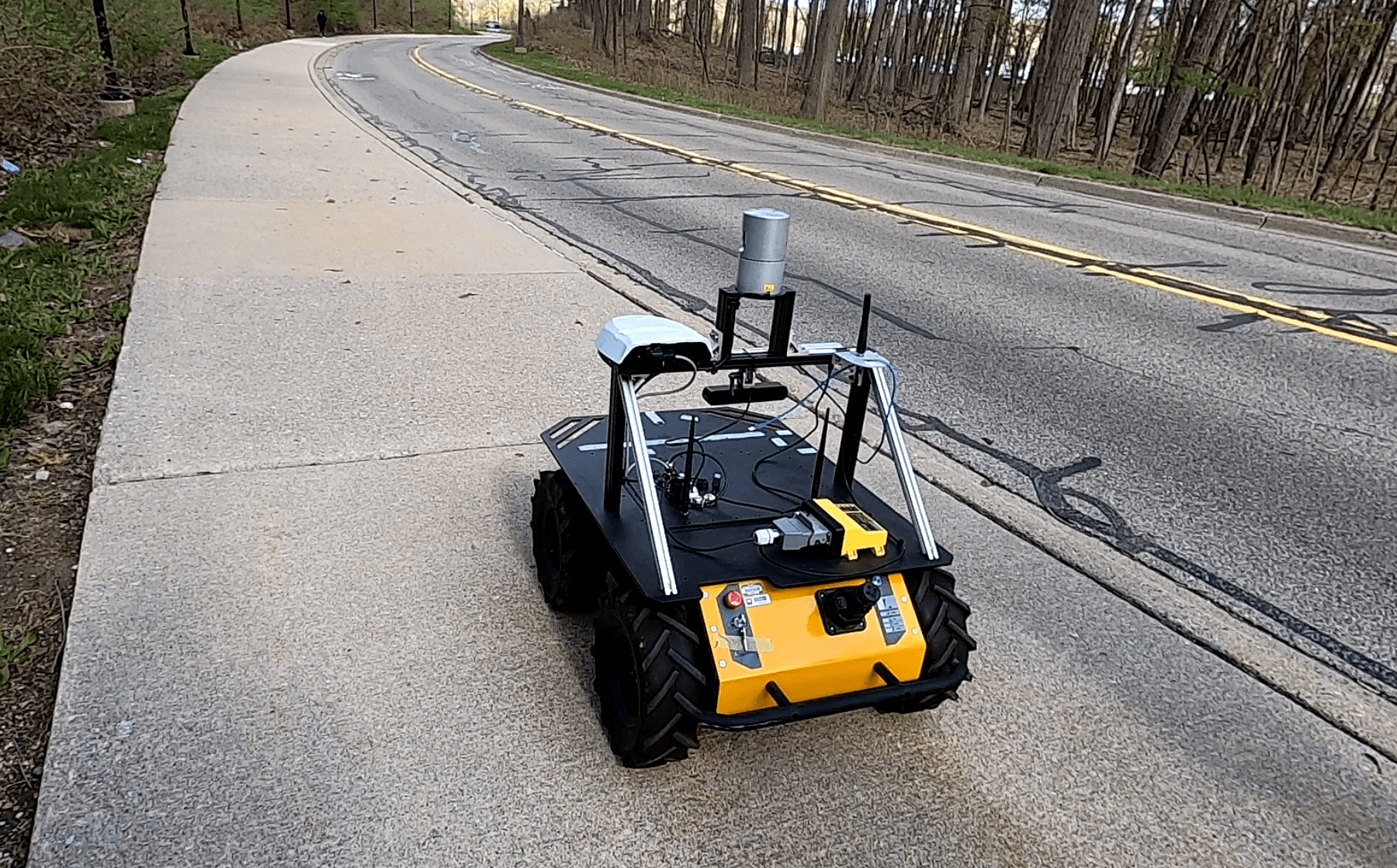}}
\caption{Estimated trajectory from DRIFT overlapped with the satellite image at the University of Michigan North Campus. A Clearpath Robotics Husky robot was driven on the sidewalk for $55$ minutes, with a total path of around 3 kilometers. Consuming proprioceptive measurements only, DRIFT can produce highly accurate estimations for long-horizon operations. This experiment demonstrates the potential of DRIFT to be a reliable odometry system in perceptually degraded situations.}
\label{fig:northwood_satellite}
\vspace{-3.5mm}
\end{figure}

To handle noisy and biased data, the fusion of different sensory data is necessary and is the standard practice in robotics~\citep{barfoot_2017}. Exteroceptive sensors such as cameras and LiDARs provide perceptual information that can assist with global localization and navigation. However, these types of sensors often operate at a low update rate (10-30 Hz) and are sensitive to perceptually degraded scenarios such as illumination changes. On the other hand, proprioceptive sensors such as Inertial Measurement Units (IMU) and joint encoders operate regardless of external factors such as lighting conditions and often at high frequencies (100-1000 $\Hz$ or higher). The high-frequency information is often important for agile navigation in challenging terrains. In addition, being agnostic to illumination changes, proprioceptive sensors can offer robust sensing modalities in extreme or featureless environments such as dense fog or sand storms. As such, designing an accurate and real-time proprioceptive state estimation algorithm is desirable. 
% This low-level state estimator can provide high-frequency pose estimations for downstream control and planning tasks. 
Contrary to Simultaneous Localization and Mapping (SLAM) systems, which focus on achieving globally consistent estimations, this proprioceptive estimator aims at obtaining high-accuracy local estimates without relying on perceptual inputs, serving as a reliable state tracking source for perception and control systems in visually challenging environments~\citep{hartley2018hybrid,ramezani2020online,teng2022error,jang2023convex}.

To develop proprioceptive state estimators, one often deals with nonlinear robot dynamics and measurement models. 
% In the past few decades, many nonlinear observers have been proposed to address this issue. Amidst them, the filtering approach is commonly preferred for real-time fusions on the resource-constraint onboard computer. 
The extended Kalman Filter (EKF)~\citep{maybeck1982stochastic,song1992extended,barfoot_2017, mcgee1985discovery} can be easily modified and has been implemented on various robotic platforms~\citep{camurri2020pronto,lee2007simulation,cobano2008location}. To address the linearization error induced in the EKF framework, many researchers have looked into leveraging symmetry~\citep{ghaffari2022progress} in the filtering framework. In particular, \citet{barrau2017invariant, barrau2018invariant} developed an invariant observer, coined Invariant extended Kalman Filter (InEKF), by modeling the state evolution on matrix Lie Groups and tracking the error on the corresponding Lie algebra. The theory of the invariant observer design is founded on estimation error being invariant under the action of a matrix Lie Group. Precisely saying, they show that if the propagation model satisfies the group affine property~\citep{barrau2017invariant}, it induces a log-linear property of the error, i.e., the nonlinear error can be exactly recovered from the time-varying linear differential equation. The InEKF has been shown to achieve better convergence and consistency in many applications, including SLAM~\citep{zhang2017convergence,barrau2015ekf,song2021right}, legged robot state estimation~\citep{hartley2020contact, lin2020contact, teng2021legged,ghaffari2022progress}, visual-inertial odometry~\citep{wu2017invariant, liu2023ingvio}, underwater navigation~\citep{potokar2021invariant} and aerial robot navigation~\citep{chen2022invariant, ko2018features}. 

Current implementations of the InEKF remain tailored for specific applications, and modifying existing libraries for different robotics applications remains cumbersome. 
% We believe the robotic community can benefit from a ready-to-use symmetry-preserving state estimation library. Therefore, 
In this work, we propose \emph{DRIFT: Dead Reckoning In Field Time}, a real-time symmetry-preserving state estimation library that is directly applicable to various robotic platforms, including legged robots, indoor and outdoor wheeled robots, full-size vehicles, and marine robots. 
% It can be used as a standalone C++ program, and we additionally provide an optional Robot Operating System (ROS) wrapper for easy communication between programs. The library can also be easily expanded with new sensor modalities. In addition, we provide two additional modules, a contact estimator and a gyro filter, to enable the proposed framework to be applied to low-cost robotic platforms. 
% % It is worth noticing that although the current implementation is based on the state-of-the-art InEKF, 
% The proposed framework can be seamlessly extended to other existing and emerging symmetry-preserving filters in the future. 

% We evaluated the proposed library on data collected from a variety of real robots, as well as simulation data for underwater robots. We showed the library achieves state-of-the-art odometry results using proprioceptive sensors only. We hope by providing a robust and accurate source of odometry, this open-sourced library can facilitate other robotic research like SLAM, control, and planning. 

\subsection{Contributions}
\label{subsec:contributions}
This work evolves from our prior work~\cite{pmlr-v164-lin22b}, in which we developed the contact estimation module for legged robot state estimation. In this work, we propose a unified framework that works for various robotic platforms. We additionally propose the gyro filter module for low-cost robots. Furthermore, we release our new real-time software, which supports various robots by default and allows easy expansion with new modalities. Lastly, we conduct extensive field experiments to assess the limits of the proposed framework on a variety of platforms, including a full-size offroad vehicle.

In particular, this work has the following contributions.
% \tyl{The first section of the main text of the journal submission (e.g., the introduction) must have a clear statement of how the submission extends or is different from the earlier conference version(s), and why the new content is significant. The earlier conference version(s) must be cited in this statement. This statement is preferably a single, standalone paragraph near the end of the first section}
\begin{enumerate}[i.]
    \item An open-source symmetry-preserving robot state estimation library that works in real-time on a variety of robotic platforms. The software can be found in \href{https://github.com/UMich-CURLY/drift}{https://github.com/UMich-CURLY/drift}.
    \item A didactic introduction to invariant Kalman filtering.
    \item A proprioceptive state estimation framework for dead reckoning that only consumes data from an onboard Inertial Measurement Unit (IMU) and kinematics for the robot, with two optional modules, a contact estimator and a gyro filter for low-cost robots. 
    \item Real-world experiments on legged robots, indoor and outdoor wheeled robots, full-size vehicles, and simulation results on marine robots to verify the proposed framework.
\end{enumerate}

% \subsection{Prior Work}
% This work evolves from our prior work~\cite{pmlr-v164-lin22b}, in which we developed the contact estimation module for legged robot state estimation. In this work, we propose a unified framework that works for various robotic platforms. We additionally propose the gyro filter module for low-cost robots. Furthermore, we release our new real-time software, which supports various robots by default and allows easy expansion with new modalities. Lastly, we conduct extensive field experiments to assess the limits of the proposed framework on a variety of platforms, including a full-size offroad vehicle.

\subsection{Outline}
\label{subsec:outline}
The remainder of this article is organized as follows. Sec.~\ref{sec:related_work} reviews the literature on different state estimation methods. Sec.~\ref{sec:preliminaries} provides some necessary backgrounds for developing an invariant filtering algorithm. Then, a didactic introduction to the invariant Kalman filtering is laid out in Sec.~\ref{sec:inekf}. Sec.~\ref{sec:drift} details DRIFT's framework and methodology. Sec.~\ref{sec:experiments} presents extensive evaluations using different robotic platforms. Limitations and future work ideas are discussed in Sec.~\ref{sec:discussion}, and Sec.~\ref{sec:conclusion} concludes the article.

\section{Literature Review}
\label{sec:related_work}
\subsection{Symmetry in Robot State Estimation}
Symmetry in Lie Groups has been explored since the 1970s~\citep {jurdjevic1972control,brockett1973lie, brockett1972system, grizzle1985structure}. Despite some early efforts on state estimation using Lie Groups~\citep{willsky1975estimation,duncan1977some}, the standard extended Kalman filter remained the mainstream method~\citep{maybeck1982stochastic, barfoot_2017, mcgee1985discovery} for decades due to its simplicity. With the standard EKF, the non-linear state dynamics and the measurement model are repeatedly linearized using the Taylor series at the current estimate~\citep{thrun2002probabilistic, barfoot_2017}. Although intuitive and easy to implement, the standard EKF can suffer from degrading performance when the dynamics become highly nonlinear, and bad initialization of states can cause the filter to diverge. In addition, because the linearization is evaluated at the current estimate, unobservable states can lead to incorrect linearization, causing spurious correlations. %Moreover, in 

% Tracking the state evolution on Euclidean spaces does not capture the "natural" evolution of the state. 
% This led to the development of symmetry-preserving observers, which capture the natural evolution of the states by modeling them on a smooth manifold (In most cases, a Lie group). 

% In the early 2000s, many researchers started to model 
The standard EKF evolves the state on Euclidean spaces; however, in robotics applications, the state often evolves on a manifold.
Alternatively, the state evolution can be modeled using Lie groups, such as quaternion~\citep{salcudean1991globally, thienel2003coupled} and $\mathrm{SO}(3)$~\citep{mahony2005complementary,hamel2006attitude}. 
% However, only a few looked into incorporating symmetry-preserving structures into the observer design. 
One of the early symmetry-preserving observers was proposed by~\citet{aghannan2003intrinsic}, where they designed an invariant observer for Lagrangian systems that is invariant under changes of the configuration coordinate. 
% Although being one of the pioneers in symmetry observer design, 
While inspiring, this method requires heavy modeling of the Lagrangian dynamics and might not be easily transferable to complex robot dynamics.

% Perhaps one key invention that enabled the development of symmetry observer design was the error-state (or indirect) extended Kalman Filter (ErEKF)~\citep{roumeliotis1999circumventing, madyastha2011extended, sola2017quaternion, trawny2005indirect}. 
Contrary to the standard EKF, which tracks the state directly, the error-state (or indirect) extended Kalman Filter (ErEKF)~\citep{roumeliotis1999circumventing, trawny2005indirect, madyastha2011extended, sola2017quaternion} tracks the state errors and their evolution. 
% One can argue that this idea of tracking the state errors partially influenced future symmetry observer design, which often tracks the evolution of the state errors on the corresponding Lie algebra. 
By tracking the error, the dynamics in ErEKF become linear under small error assumptions~\citep{roumeliotis1999circumventing}. However, despite the near-linear dynamics, the ErEKF still parameterizes the state in Euclidean spaces. To be more specific, the attitude is often represented using Euler angles. These representations contain singularities (i.e., do not cover the entire rotation manifold), and the filter can consequently be trapped in the famous Gimbal lock. The multiplicative (or quaternion) EKF (MEKF)~\citep{markley2003attitude, sola2017quaternion, trawny2005indirect} addresses this problem by tracking the rotation using quaternions and updating the state using multiplicative actions. The orientation error is modeled as a $3$-vector on the tangent space of $\mathrm{SO}(3)$ to maintain the nonsingular covariance~\cite{trawny2005indirect}. 
% This is often achieved using small error assumptions~\citep{sola2017quaternion, trawny2005indirect}. 
The MEKF has been successfully implemented on many platforms, including aerial vehicles~\citep{koch2020relative, hall2008quaternion,leishman2015multiplicative}, legged robots~\citep{bloesch2013state, rotella2014state}, and marine robots~\citep{bonin2013multisensor}. Nevertheless, the linearized error propagation of the MEKF still depends on the estimated states, which can degrade the filter performance. 

In invariant observer design, the estimated error is invariant under the action of a matrix Lie group~\citep{bonnabel2009non}. This resulted in developing the InEKF~\citep{bonnabel2007left,barrau2015non,barrau2017invariant,barrau2018invariant} where the state is represented using a matrix Lie group, and the error is tracked in the corresponding Lie algebra. If the system satisfies the ``group-affine'' property, the estimation error follows a ``log-linear'' dynamics in the Lie algebra~\citep{barrau2015non,barrau2017invariant,barrau2018invariant}. Namely, the system linearization is independent of the state estimates. Because of this invariance property, the InEKF can achieve stronger convergence on many state estimation applications, including SLAM~\citep{zhang2017convergence,barrau2015ekf,song2021right}, legged robot~\citep{hartley2020contact, lin2020contact, teng2021legged, gao2022invariant}, visual-inertial odometry~\citep{wu2017invariant, liu2023ingvio}, underwater robots~\citep{potokar2021invariant} and aerial vehicles~\citep{chen2022invariant, ko2018features, barczyk2012invariant}. In addition to filtering, some researchers also investigated the potential of symmetry-preserving structures in optimization-based smoothing frameworks~\citep{chauchat2018invariant,chauchat2022invariant, kim2021legged, huai2021consistent}. 
% Contrary to the filtering approach, which marginalized the past information, the smoothing framework retains the history and optimizes the states jointly (often by solving a maximum a posteriori (MAP) problem). Using the same theorem from the InEKF, the linearized Jacobian becomes state-invariant if the system dynamics satisfy the group-affine property. 
The invariant smoother framework can perform better than InEKF at the cost of higher computational time~\citep{chauchat2018invariant,kim2021legged}.

More recently, a generalization of the InEKF framework, the Equivariant Filter (EqF), has been proposed by~\citet{van2020equivariant} and~\citet{mahony2021equivariant}. In EqF, the state evolves on a homogeneous space (smooth manifolds with transitive Lie group actions). By finding an appropriate lift, the system dynamics can be realized on symmetry Lie groups. The correction, on the other hand, is derived by linearizing the error kinematics with respect to a fixed origin on the homogeneous space~\citep{van2020equivariant,mahony2021equivariant,van2023equivariant}. Because the EqF does not require the system to be modeled explicitly on Lie groups, it relaxes the group-affine constraint posed in the InEKF and works with any system with equivariant properties. In fact, it can be reduced to the InEKF when the corresponding state evolves on matrix Lie groups and satisfies the group-affine property~\citep{mahony2021equivariant, van2023equivariant}. Moreover, since the state is modeled on homogeneous spaces, EqF opens up the opportunity to estimate the input bias without breaking the symmetry of the system~\citep{fornasier2022overcoming,fornasier2022equivariant}, which is not feasible in the invariant framework~\cite{barrau2018invariant,hartley2020contact}. The EqF demonstrates great potential in different applications~\citep{van2021equivariant,van2022eqvio,luo2021equivariant} and can be integrated into DRIFT in the future.

% Using quaternion,~\citet{lefferts1982kalman, markley2003attitude} proposed the Multiplicative Extended Kalman Filter (MEKF), where the estimation error was modeled as multiplicative to the quaternion state instead of the traditional addictive approach. 

% The invariant property was first explored in the 2000s by~\citet{aghannan2003intrinsic}, where they designed an invariant observer for Lagrangian systems. 

\subsection{Legged Robots}
% Legged robots are a special subclass of mobile robots that move forward by alternating leg contact points in the environment. Because of the unstable nature of such motion, 
The controller of legged robots often requires accurate and high-frequency state estimations to maintain the robot's stability. This can be achieved by fusing multiple onboard sensors such as IMU, joint encoders, and cameras.
% In this subsection, we review some important legged robot state estimation algorithms.

\subsubsection{State Estimation}
One simple way to obtain the state of a legged robot is through leg kinematics, known as leg odometry~\citep{roston1991dead}. By assuming the contact foot to be static in the world frame, one can infer the robot's state using forward kinematic functions and encoder measurements. Leg odometry is simple and has been successfully implemented on many early legged robotic platforms, including CMU Amhler~\citep{roston1991dead}, RHex hexapod~\citep{lin2005leg}, and MARLO~\citep{da20162d}. However, it often performs poorly due to noisy encoder data, inaccuracies in kinematic modeling, and foot slip~\citep{roston1991dead}. This results in noisy velocity and unbounded position and orientation estimates. To address this, \citet{lin2005leg} used prior terrain information and three non-colinear contact feet to estimate the instantaneous robot's pose through kinematics. Although achieving better accuracy, this algorithm cannot operate when two or more feet are in the aerial phase. As a result, it significantly hampers any agile movements of the robot. 

% In addition to encoders on the legs, 
Modern legged robots are often equipped with extra sensors like IMU, camera, and LiDAR. 
% This gives researchers a means to improve the state estimation accuracy by fusing multiple sensors with the leg kinematic measurements. In their following work,~
\citet{lin2006sensor} fused leg kinematics with IMU measurements using an EKF for the RHex hexapod. They showed that by fusing the two measurements, the algorithm outperforms methods that only use one modality in isolation. In addition to IMU and leg odometry, researchers also explored fusing different sensor modalities within an EKF framework, including fusing leg odometry with a magnetometer and GPS~\citep{cobano2008location}, and leg odometry with optical flow~\citep{singh2006optical}. By fusing multiple sensor measurements, the above methods improve the estimation accuracy of pure leg odometry. However, they still require heavy modeling of the robot's dynamics, which can be cumbersome and inaccurate. 

\citet{bloesch2012state} proposed observability-constrained ErEKF to fuse IMU and kinematic measurements. In this framework, the process model does not depend on the robot or the gait motion. Instead, the readings from an inertial sensor are integrated to obtain the prediction (also known as the IMU strapdown modeling). 
% To elaborate, the angular velocity and linear acceleration readings are integrated to propagate the robot's motion, and the covariance is propagated by tracking the error dynamics of the states. 
Hence, the need for complicated dynamic modeling is circumvented~\citep{roumeliotis1999circumventing}. Another innovation of this work is the state augmentation of the contact foot position. 
% This is achieved by augmenting contact foot position into the states when the foot is first in contact with the ground and marginalizing it out when the contact breaks. Similar to the leg odometry~\citep{roston1991dead}, the contact foot position is assumed to be static during the contact phase. As a result, the dynamics of the new augmented contact state during this period are assumed to be constant, with noises to compensate for foot slips. 
This augmented position serves as a fixed reference position in the world frame. Therefore, the mismatch between new kinematic measurements and the predicted contact state can then be used as the correction model. 
% With IMU strapdown modeling and contact augmentation, it opens up a door for general robot state estimator designs that work on any legged robot. In fact, 
IMU strapdown has become a common approach for modern state estimators~\citep{fallon2014drift,bloesch2013state,xinjilefu2014decoupled,hartley2018hybrid}. \citet{rotella2014state} includes the foot orientation measurements from forward kinematics in an ErEKF for humanoid robots. However, this method only works when the foot does not rotate during the contact phase. 

% In addition to the above filtering approaches, some researchers address the problem using optimization-based smoothing methods. These types of methods often retain the history of the states and measurements in a factor graph formulation, and jointly optimize them using non-linear optimization algorithms. 
In the context of optimization-based methods instead of filtering, \citet{hartley2018legged} proposed a factor graph formulation to fuse leg odometry with IMU measurements on a bipedal Cassie robot. They introduced new forward kinematics and pre-integrated contact factors to include additional constraints posed by the non-slip contact assumption. \citet{hartley2018hybrid} further incorporated visual information from semi-direct visual odometry (SVO)~\citep{forster2016svo} to form a visual-inertial-legged smoother. Instead of using SVO as an additional module, \citet{wisth2019robust} proposed VILENS (Visual Inertial LEgged Navigation System), which tightly integrates visual features from an RGB-D camera into the cost function. VILENS is implemented on an ANYmal robot and showed improved performance over previous methods. Built upon VILENS, the same group introduced the foot velocity integration factor in~\citep{wisth2020preintegrated}. By integrating the foot velocity instead of position, a slow-varying velocity bias can be inferred from vision to account for foot slip events. \citet{kim2022step} exploited the body velocity estimation from a camera and dropped the need for the contact estimation module. Their method was implemented on an MIT mini cheetah and shown to achieve superior performance on slippery terrains. When only proprioceptive sensors were presented,~\citet{kim2021legged} proposed a slip-rejection mechanism in addition to a proprioceptive smoother framework using an MIT mini cheetah. This method improved the robustness of state estimation even in the absence of camera information. Smoothing approaches have the privilege of leveraging past information during optimization. However, this comes at the cost of higher computational efficiency than filtering, especially during long-horizon experiments, where the past factors could grow unboundedly. Moreover, the above methods did not take advantage of the existing symmetry structure present in the problem, which improves the performance and consistency of the estimator.

% More recently, researchers started to incorporate symmetry-preserving structures in legged robot observer design. 
\citet{hartley2020contact} introduced the InEKF to the legged robot community by designing a contact-aided InEKF filtering. Similar to the work of~\citet{bloesch2012state}, the IMU strapdown model was used for propagation, and the contact states were augmented for correction. However, in this work, the states were tracked on the Lie group $\mathrm{SE}_k(3)$~\citep{barrau2015non}, and the error dynamics evolved in the corresponding Lie algebra. \citet{hartley2020contact} showed that the contact-inertial process model satisfies the group-affine property~\citep{barrau2018invariant}, and the resulting linearization matrix is independent of the estimated state. The contact-aided InEKF demonstrated state-of-the-art performance on a bipedal Cassie robot and has since been modified for different applications~\citep{gao2021invariant,gao2022invariant,lin2020contact,teng2021legged}. 

% In the past, several researchers have released open-sourced legged robot state estimators. 
Existing open-sourced libraries legged robot state estimation, include 
% can facilitate other robotic research and are greatly beneficial to the community. One example is 
Pronto proposed by~\citet{camurri2020pronto}, EKF-based framework for both exteroceptive and proprioceptive sensor fusion. The framework supports multiple bipedal and quadruped robots. However, Pronto is based on the traditional EKF and the state-dependent linearization could lead to sub-optimal results. Another example is Cerberus proposed by~\citet{yang2023cerberus}, which is a factor graph framework for visual-inertial-leg odometry. One key feature of Cerberus is the ability to estimate kinematic parameters online. However, Cerberus relies on visual information for the kinematic parameters estimation, which can lead to degraded performance in visually challenging environments. Moreover, Cerberus is reported to operate at $20 \Hz$, which might be insufficient for some control algorithms. In addition to the above two libraries, \citet{hartley2020contact} also released their source code for contact-aided InEKF for Cassie. In DRIFT, we adopt the software by \citet{hartley2020contact}, and our previous work~\citep{lin2020contact}, and enable modular extensions to different modalities. 

\subsubsection{Contact Estimation}
Reliable contact estimation that accurately captures the zero velocity events on foot is of crucial importance for legged robot state estimators. 
% Below we briefly review some contact estimation methods in the literature.
Model-based approaches segment the touchdown event of robot legs or prosthetic legs by thresholding of
the estimated Ground Reaction Force (GRF) from the general equation of motion~\citep{focchi2013local,fakoorian2016ground,fink2020proprioceptive}. Although this method can detect touchdown events, the estimated GRF is often noisy and unreliable, especially for lightweight robots. \citet{de2006collision, haddadin2008collision} proposed a Generalized Momentum (GM) method for detecting contact events on robot manipulators. This GM-based method is, in fact, a filtered version of the work of~\citet{focchi2013local}. Although GM-based methods mitigate the noise problem in GRF estimation, an empirical threshold on the cut-off frequency is still required. 

\citet{hwangbo2016probabilistic} used a probabilistic representation of the contact state and a Hidden Markov Model (HMM) to fuse the dynamics and kinematics for contact estimation by adopting a Monte-Carlo sampling algorithm to compute the transition model and verifying against GM-based methods. \citet{jenelten2019dynamic} expanded the HMM method and focused on slippage detection. They demonstrate ANYmal~\citep{hutter2016anymal}, a quadruped robot, walking stably on slippery ground. The above two methods aim to detect contact as early as possible for the controller to maintain stability; however, we aim to detect contact intervals for state estimation on various terrains. 

\citet{bledt2018contact} leveraged the GM-based methods and the probabilistic representation of contact states by using a Kalman filter to fuse the gait phase scheduler information from the controller with the GRF 
% estimated using the GM method 
and demonstrate that estimated contacts can assist the controller in reducing the bouncing event upon touchdown. However, this method assumes the leg phase to be periodic as it uses the gait scheduler information in the prediction step of the Kalman filter. It could experience a loss in performance when the phase is heavily violated as the robot interacts with uneven terrains. 

% Data-driven approaches take advantage of the rapid development of recent machine learning techniques. 
\citet{camurri2017probabilistic} used logistic regression to learn the GRF threshold for contact detection. This work compares against heuristic-based thresholding on GRF using a base state estimator. The result shows that the logistic regression classifier can double the performance of the state estimator. However, compared to deep learning methods, the performance of a logistic regression classifier gets saturated as the amount of data increases~\citep{lecun2015deep,Goodfellow-et-al-2016}.
Moreover, this method requires a specific training procedure for different gait, loading conditions of the robot, and individual terrain properties.  

\citet{rotella2018unsupervised} used a fuzzy C-means clustering for the probability of contacts in all six end-effector degrees of freedom. They integrate the contact estimator with a base state estimator and show their approach performs considerably better than implementations purely based on measured normal force. However, this method assumes contact wrench sensors and additional IMU are available at each end-effector. Furthermore, this method was only tested in simulation. Its performance on real robots remains unknown. \citet{piperakis2019unsupervised} proposed an unsupervised learning method for humanoid gait phase estimation by employing Gaussian Mixture Models (GMMs) for clustering and comparing to the ground-truth data and leg odometry. However, this work also assumes the availability of wrench/force sensors at each end-effector, and the clustering result is affected by the gait and data density. 

The above methods either assume the availability of wrench/force sensors or are restricted by the nature of simple regression and are thus unable to generalize to different scenarios. In contrast, our work proposes a multi-modal deep learning-based contact estimator that does not require contact sensors and can generalize well to different gaits and terrain properties. Moreover, as more data becomes available, the network performance can be improved. 

\subsection{Wheeled Robots}
% Citep and citet?
Wheeled robots and autonomous vehicles state estimation problems often treat odometer measurements and dynamics constraints as given, which increases the accuracy and robustness of the localization systems~\citep{wu2017vins, chae2020robust}. However, such methods require high computational power to process image flow. The IMU-based algorithms are more universal and approachable for wheeled robots, which can contribute to emergency events detection such as collision and slippage~\citep{yu2022fully, xiong2019imu}. 

For outdoor wheeled robots, GPS plays an important role in exteroceptive localization which uses multiple satellites to obtain the robot's position in the earth coordinates. Nevertheless, the GPS accuracy is limited by radio signal transmission time, and the slow-down effect of water vapor and other particles in the atmosphere counts for the propagation delay~\citep{bajaj2002gps}. The occlusions and reflections of and between the buildings, namely multipath fading, will also reduce the GPS localization accuracy. To overcome these limitations, beacon-based methods, such as RFID~\citep{hahnel2004mapping, chae2005combination}, QR-code~\citep{zhang2015localization, lee2014autonomous}, WiFi or Bluetooth~\citep{yang2015wifi, lashkari2010wifi,jadidi2017gaussian,ghaffari2018radio}, and UWB~\citep{zhang2006accurate, kuhn2010adaptive}, are used for urban and indoor environments localization.

In terms of proprioceptive state estimation for four-wheeled chassis robots, typically autonomous vehicles, the measurements are IMU readings, steering wheel angle, and wheel angular velocity, which is the most common sensor configuration for vehicle velocity estimation~\citep{imsland2006vehicle, imsland2007nonlinear}. Nonlinear observer (NLO), nonlinear unknown input observer (NUIO), and reduced nonlinear observer (RNLO) are proposed for vehicle lateral and longitudinal velocity estimation by~\citet{imsland2006vehicle, imsland2007non, guo2013design}. \citet{wenzel2006dual} invented the dual extended Kalman filter (DEKF) for combined estimation of vehicle states and vehicle parameters. 

Mobile robot filtering-based localization performance can be increased by augmenting uncertain parameters into the state~\citep{durrant1996autonomous}. Nevertheless, with model-based filters such as the Kalman filter, the performance highly depends on how accurately the stochastic model can capture the underlying physical process. \citet{kwon2006effective} proposed a combined Kalman filter-perturbation estimator (CKF) for wheeled robots, which does not require any uncertainty model except for the noise statistics. The perturbation estimator generates equivalent perturbations based on the nominal state equation and action model of the robot, which is simultaneously corrected by the estimates of perturbation. The perturbation estimator's integral control module leads to a decrease in the localization error. 

A dead-reckoning system has an unavoidable drift over time because, first, the integration of sensors' random walk and, secondly, inaccuracies of the system modeling for the action and observation models. \citet{welte2019four} presented a dead-reckoning model that can fuse all common sensors and then merge the redundant ones, increasing the robustness compared with the traditional Zero Velocity Update (ZUPT) method. The work also proposed a method to compensate for the systematic sensor and observation model errors, which relies on the Rauch-Tung-Striebel smoothing method to estimate the robot state and further calibrate the system model parameters. 

\subsection{Marine Robots}
Underwater environments are often featureless with limited visibility, and objects underwater are frequently subject to unexpected external forces from ocean currents or surges. This poses extra challenges for Autonomous Underwater Vehicles (AUVs) state estimation. To make matters worse, some common sensors for ground robots, such as joint encoders and GPS, cannot be directly applied to AUVs. However, many widely used modern underwater localization systems depend on accurate dead reckoning solutions~\citep{paull2013auv}.
% As a result, special algorithms need to be developed to overcome these issues. 
% Fortunately, most AUVs are equipped with sensors that are particularly suited for underwater sensing. 
% , which we will discuss in the next paragraph. 
% Due to these unique challenges and different sensor modalities, traditional dead reckonings are not considered a primary means for underwater navigation.  Therefore, 
\emph{We position dead reckoning algorithms such as DRIFT as a submodule of a more complex localization system.} Comprehensive reviews on underwater state estimators can be found in~\citep{kinsey2006survey, paull2013auv,  maurelli2021auv}. 
% Below we briefly review some commonly used underwater sensors and related state estimation algorithms.

There are several specialized sensors for underwater navigation. Because the water pressure can roughly translate to the depth below sea level, a pressure sensor can be used to obtain depth measurements. In addition to pressure sensing, acoustic sensors such as Doppler Velocity Logs (DVL)~\citep{rudolph2012doppler} also play a big role in underwater navigation. A DVL can provide ground-referenced velocity measurements. This is achieved by emitting directional acoustic beams to the seabed and using the Doppler effect on the reflected acoustic signals. Unlike the above sensors that are mounted directly on the robot, transponder systems~\citep{milne1983underwater}, such as a Long Baseline system (LBL), use floating beacons to provide positioning measurements from triangulation. Although transponder systems allow direct positioning measurements, they require additional beacons to be deployed, limiting the operation range of AUVs. 

% To obtain accurate state estimation results, fusions of the above sensors are needed. 
% Similar to ground robots, 
EKF has been implemented on multiple AUVs due to its simplicity~\citep{lee2007simulation, huang2010autonomous, alcocer2007study}. \citet{karras2010line} proposed a Dual Unscented Kalman Filter (DUKF) framework to fuse measurements from a laser sensor and an IMU. The filter jointly estimated the 6D pose of the robot and its dynamic parameters. 
% This method was evaluated in two control schemes and was shown to have improved performance compared to the classical UKF. Nonetheless, 
The DUKF was still sensitive to initialization, and a specific closed-loop control scheme was required during initialization. Instead of online optimization of the dynamic parameters, \citet{hegrenaes2011model} formulated a more accurate vehicle model in a model-aided EKF framework. The kinetic model was integrated to provide velocity measurements for filter correction. 
% In order to account for unexpected sea currents in the kinetic model, 
Sea current velocity was treated as a slowly varying bias 
% to the water relative velocity, 
and was estimated along with the robot's state. The method was evaluated using a HUGIN 4500 AUV on various field tests. The IMU strapdown modeling was also adopted for marine robotic applications to circumvent complex modeling of the vehicle. Using an ErEKF framework, ~\citet{miller2010autonomous} fused multiple sensors, including an IMU, an LBL, a DVL, a pressure sensor, and an attitude sensor for UAV state estimation. Additional calibration parameters, such as the speed of sound in seawater, were also estimated to reduce systematic errors. This framework was shown to be robust to sporadic sensor failures. However, a prolonged failure of the DVL or LBL can degrade the performance, and the requirement of LBL still limits the operation range. %robots from achieving full autonomy.  

% Surprisingly, symmetry-preserving state estimators have not yet become popular for marine robotic applications. To the best of our knowledge, the only symmetry-preserving state estimator for AUVs is the work proposed by~\citet{potokar2021invariant}. In this work, 
\citet{potokar2021invariant} used an InEKF to estimate the 6D pose of an AUV. The velocity readings from a DVL, and depth measurements from a pressure sensor were fused with the IMU strapdown model. Because the depth measurements do not obey the left-invariant formulation, which we will discuss later in Sec.~\ref{sec:preliminaries}, pseudo measurements with infinity covariances were introduced to allow the incorporation of such measurements. This framework was evaluated in a simulation environment and was shown to have better convergence over the MEKF. The algorithm was implemented in Python and was demonstrated to have real-time performance. 
% However, it was difficult to expand the framework with new modalities and it did not support standard communication packages such as ROS. On the other hand, 
DRIFT is a unified state estimator that directly supports multiple robots. Modular implementation in C++ enables faster computation on resourced-constrained devices and allows easy expansion with new modalities. In addition, for convenience, we provide a ROS wrapper for real-time communication on robots.

\section{Preliminaries and notation}
\label{sec:preliminaries}
Matrix Lie groups~\citep{chirikjian2011stochastic,hall2015lie,barfoot_2017} provide natural (exponential) coordinates that exploits symmetries of the space~\citep{long2013banana,barfoot2014associating,forster2016manifold,mangelson2020characterizing,mahony2021equivariant,brossard2021associating}. State estimation is the problem of determining a robot's position, orientation, and velocity that are vital for robot control~\citep{barfoot_2017}. An interesting class of state estimators that can be run at high frequency, e.g., 2 $\mathrm{kHz}$, are based on InEKF~\citep{barrau2015non,barrau2017invariant,barrau2018invariant,hartley2020contact}. The theory of invariant observer design is based on the estimation error being invariant under the action of a matrix Lie group. The fundamental result is that by correct parametrization of the error variable, a wide range of nonlinear problems can lead to log-linear error dynamics~\citep{bonnabel2009non,barrau2015non,barrau2017invariant}.

Consider a deterministic Linear Time-Invariant (LTI) process model
% \begin{equation}
% \label{eq:lti_sys}
 $\dot{x} = Ax+Bu$.    
% \end{equation}
Let $\bar{x}$ be an estimate of $x$, i.e., $\dot{\bar{x}} = A\bar{x}+Bu$. Define the error $e := x - \bar{x}$. Then $\dot{e} = \dot{x} - \dot{\bar{x}} = A (x - \bar{x}) = Ae$ is an autonomous differential equation. Given an initial condition $e(0) = e_0$, we can solve for the error at any time via $e(t) = \exp(At) e_0$. 
In other words, error propagation is independent of the system trajectory, i.e., state estimate. The Invariant EKF~\cite{barrau2017invariant} generalizes this observation to linear-exponential systems, e.g., rigid body systems, using the framework of matrix Lie groups.

\subsection{Process Dynamics on Lie Groups}
\label{subsec:lie-process}

A process dynamics evolving on the Lie group $\mathcal{G}$, for state $X_t \in \mathcal{G}$, is
 \begin{equation}
  \frac{d}{dt} X_t = f_{u_t}(X_t).
 \end{equation}
$\bar{X}_t$ denotes an estimate of the state. The state estimation error is defined using right or left multiplication of $X_t^{-1}$.

\begin{definition}[Left and Right Invariant Error] 
The right- and left-invariant errors between two trajectories $X_t$ and $\bar{X}_t$ are:
\begin{equation} \label{eq:invariant_error}
\begin{split}
\eta_t^r &= \bar{X}_t X_t^{-1} = (\bar{X}_t L) (X_t L)^{-1} \quad \text{(Right-Invariant)}\\
\eta_t^l &= X_t^{-1} \bar{X}_t = (L \bar{X}_t)^{-1} (L X_t), \quad \text{(Left-Invariant)}
\end{split}
\end{equation}
where $L \in \mathcal{G}$ is an arbitrary element of the group.
\end{definition}

\begin{theorem}[Autonomous Error Dynamics~\citep{barrau2017invariant}] \label{theorem:autonomous_error_dynamics}
A system is group affine if the dynamics, $f_{u_t}(\cdot)$, satisfies:
\begin{equation} 
\label{eq:group_affine}
f_{u_t}(X_1 X_2) = f_{u_t}(X_1) X_2 + X_1 f_{u_t}(X_2) - X_1 f_{u_t}(I) X_2
\end{equation}
for all $t>0$ and $X_1, X_2 \in \mathcal{G}$. Furthermore, if this condition is satisfied, the right- and left-invariant error dynamics are trajectory-independent and satisfy:
\begin{align}
\label{eq:error_dynamics}
\nonumber \frac{d}{dt} \eta_t^r &= g_{u_t}(\eta_t^r) \quad \text{where} \quad
g_{u_t}(\eta^r) = f_{u_t}(\eta^r) - \eta^r f_{u_t}(I) , \\
\frac{d}{dt} \eta_t^l &= g_{u_t}(\eta_t^l) \quad \text{where} \quad
g_{u_t}(\eta^l) =  f_{u_t}(\eta^l) - f_{u_t}(I) \eta^l .
\end{align}
\end{theorem}

$I$ denotes the group identity element ($I_n$ for $n\times n$ identity matrix). Define $A_t$ to be a $\mathrm{dim} \mathfrak{g} \times \mathrm{dim} \mathfrak{g}$ matrix, where $\mathfrak{g}$ is the Lie algebra (tangent space at the identity) of $\mathcal{G}$, satisfying 
\begin{equation}
\label{eq:linear_approx_dynamic}
    g_{u_t}(\exp(\xi)) := (A_t \xi)^\wedge + \mathcal{O}(\lVert \xi \rVert^2).
\end{equation}
For all $t \ge 0$, let $\xi_t \in \mathbb{R}^{\mathrm{dim} \mathfrak{g}}$ be the solution of  the linear differential equation $\frac{d}{dt} \xi_t = A_t \xi_t$.
% \pause

 \begin{theorem}[Log-Linear Property of the Error~\citep{barrau2017invariant}] \label{theorem:log_linear_error}
Consider the right-invariant error, $\eta_t$, between two trajectories (possibly far apart). For arbitrary initial error $\xi_0 \in \mathbb{R}^{\mathrm{dim} \mathfrak{g}}$, if 
\mbox{$\eta_0 =\exp(\xi_0)$}, then for all $t\ge 0$, 
% \begin{equation*}
$\eta_t = \exp(\xi_t)$;
% \end{equation*}
that is, the nonlinear estimation error $\eta_t$ can be exactly recovered from the time-varying linear differential equation.
\end{theorem}

For an example of the difference with Euler angle parametrization, see~\citet[Sec. 4]{hartley2020contact}.

\subsection{Associated Noisy System}

A noisy process dynamics evolving on the Lie group take the following form.
\begin{equation*} 
\frac{d}{dt} X_t = f_{u_t}(X_t) + X_t w_t^\wedge ,
\end{equation*}
 where $w_t^\wedge \in \mathfrak{g}$ is a continuous white noise whose covariance matrix is denoted by $Q_t$. Then the equivalent noisy error dynamics can be shown to be
 \begin{align}
 \label{eq:noisyerrdyn}
     \nonumber \frac{d}{dt} \eta_t^r &= g_{u_t}(\eta_t^r) - (\bar{X}_t w_t^\wedge \bar{X}_t^{-1}) \eta_t^r \\
     \nonumber &= g_{u_t}(\eta_t^r) - (\mathrm{Ad}_{\bar{X}_t} w_t)^\wedge \eta_t^r, \\
     \frac{d}{dt} \eta_t^l &= g_{u_t}(\eta_t^l) - w_t^\wedge \eta_t^l ,
 \end{align}
 where for every $X \in \mathcal{G}$, the adjoint action, $\mathrm{Ad}_{X}: \mathfrak{g}\rightarrow \mathfrak{g}$, is a Lie algebra isomorphism that enables change of frames (matrix similarity).

\subsection{Invariant Observation Model}
During the correction step of a Kalman filter, the error is updated using incoming sensor measurements. If observations take a particular form, then the linearized observation model and the innovation will also be autonomous. This happens when the measurement, $Y_{t_k}$, can be written as either
\begin{equation} 
\label{eq:invariant_observations}
\begin{alignedat}{2} 
Y_{t_k} &= X_{t_k} b + V_{t_k}  \quad &&\text{(Left-Invariant Observation)}~~~\text{or} \\ 
Y_{t_k} &= X_{t_k}^{-1} b + V_{t_k} \quad &&\text{(Right-Invariant Observation)}.
\end{alignedat}
\end{equation}
$b$ is a constant vector and $V_{t_k}$ is a vector of Gaussian noise.

\section{Invariant Kalman Filtering}
\label{sec:inekf}
\subsection{Left-Invariant Extended Kalman Filter}
\label{sec:li-ekf}
We now derive the Left-Invariant Extended Kalman Filter (LI-EKF). 
% following the Kalman filtering theory in the previous chapters. 
In particular, the filter tracks the state mean and covariance as parameters and corrects the error via a ``linear'' update rule. As we will see, the innovation is in the Lie algebra, and the update rule follows the group operation for integration.

\subsubsection{Propagation}
In the propagation (or prediction) step, the mean can be directly computed using the process model. However, for the covariance propagation, we use the left-invariant log-linear error dynamics. As such, the state's mean evolves on the group while the covariance is tracked in the Lie algebra. 
    \begin{align} 
        \nonumber \frac{d}{dt} \bar{X}_t &= f_{u_t}(\bar{X}_t), \quad t_{k-1} \leq t < t_k, \\
    % \end{equation*}
    % \begin{equation*} 
        \nonumber \frac{d}{dt} \eta_t^l &= g_{u_t}(\eta_t^l) - w_t^\wedge \eta_t^l \implies \frac{d}{dt} \xi_t^l = A_t^l \xi_t^l - w_t, \\
    % \end{equation*}
    % \begin{equation*} 
        \frac{d}{dt} P_t^l &= A_t^l P_t^l + P_t^l {A_t^l}^\transpose + Q_t .
    \end{align}

\subsubsection{Update}
In the update (or correction) step, we use \mbox{$\eta = \exp(\xi) \approx I + \xi^\wedge$} and neglect the higher order terms to derive the linear update error equation as follows.
    \begin{align} 
        \nonumber \bar{X}_{t_k}^+ &= \bar{X}_{t_k} \exp \left( L_{t_k} \left( \bar{X}_{t_k}^{-1} Y_{t_k} - b \right) \right), \\
    % \end{equation*}
    % \begin{equation*} 
        \nonumber X_{t_k}^{-1} \bar{X}_{t_k}^+ &= X_{t_k}^{-1} \bar{X}_{t_k} \exp \left( L_{t_k} \left( \bar{X}_{t_k}^{-1} (X_{t_k} b + V_{t_k}) - b \right) \right) ,\\
    % \end{equation*}
    % \begin{equation*} 
        \nonumber \eta_{t_k}^{l+} &= \eta_{t_k}^{l} \exp \left( L_{t_k} \left( (\eta_{t_k}^{l})^{-1} b - b + \bar{X}_{t_k}^{-1} V_t \right) \right) ,\\
    % \end{equation*}
    % \begin{equation*} 
    %     I + {\xi_{t_k}^{l+}}^\wedge = (I + {\xi_{t_k}^{l}}^\wedge) \left(I + \left( L_{t_k} \left( (I - {\xi_{t_k}^{l}}^\wedge) b - b + \bar{X}_{t_k}^{-1} V_t \right) \right)^\wedge \right)
    % \end{equation*}
    % \begin{equation*} 
            \nonumber {\xi_{t_k}^{l+}}^\wedge &= {\xi_{t_k}^{l}}^\wedge + \left( L_{t_k} \left( (I - {\xi_{t_k}^{l}}^\wedge) b - b + \bar{X}_{t_k}^{-1} V_t \right) \right)^\wedge , \\
    % \end{equation*}
    % \begin{equation*} 
            \xi_{t_k}^{l+} &= \xi_{t_k}^{l} + L_{t_k} \left( -{\xi_{t_k}^{l}}^\wedge b + \bar{X}_{t_k}^{-1} V_t \right).
    \end{align}

To arrange all terms according to the vector form of $\xi$, define the measurement Jacobian, $H$, such that 
$\boxed{H \xi = \xi^\wedge b} .$
Then it follows that
    \begin{align} 
        \nonumber \bar{X}_{t_k}^+ &= \bar{X}_{t_k} \exp \left( L_{t_k} \left( \bar{X}_{t_k}^{-1} Y_{t_k} - b \right) \right), \\
    % \end{equation*}
    % \begin{equation*} 
    %         \xi_{t_k}^{l+} = \xi_{t_k}^{l} - L_{t_k} H \xi_{t_k}^{l} + L_{t_k} \bar{X}_{t_k}^{-1} V_t
    % \end{equation*}
    % \begin{equation*} 
           \nonumber  \xi_{t_k}^{l+} &= (I - L_{t_k} H ) \xi_{t_k}^{l} + L_{t_k} \bar{X}_{t_k}^{-1} V_t, \\
    % \end{equation*}
    % \begin{equation*}
        P_{t_k}^{l+} &= (I - L_{t_k} H) P_{t_k}^l (I - L_{t_k} H)^\transpose + L_{t_k} \bar{N}_k L_{t_k}^\transpose, 
    \end{align}
    where
    % \begin{equation*}
        $\bar{N}_k := \bar{X}_{t_k}^{-1} \operatorname{Cov}(V_k) \bar{X}_{t_k}^{-\transpose}$.
    % \end{equation*}

\subsubsection{LI-EKF Result}
To summarize, we have the following two steps (as usual for a Kalman filter).
\begin{mdframed}
\begin{enumerate}
  \item LI-EKF Propagation:
    \begin{align} 
        \nonumber  \frac{d}{dt} \bar{X}_t &= f_{u_t}(\bar{X}_t), \quad t_{k-1} \leq t < t_k, \\
    % \end{equation}
    % \begin{equation} 
        \frac{d}{dt} P_t^l &= A_t^l P_t^l + P_t^l {A_t^l}^\transpose + Q_t.
    \end{align}
  \item LI-EKF Update: 
    \begin{align} 
        \nonumber  \bar{X}_{t_k}^+ &= \bar{X}_{t_k} \exp \left( L_{t_k} \left( \bar{X}_{t_k}^{-1} Y_{t_k} - b \right) \right), \\
    % \end{equation}
    % \begin{equation}
        P_{t_k}^{l+} &= (I - L_{t_k} H) P_{t_k}^l (I - L_{t_k} H)^\transpose + L_{t_k} \bar{N}_k L_{t_k}^\transpose, 
    \end{align}
    where 
    \begin{equation}
        L_{t_k} = P_{t_k}^l H^\transpose S^{-1}, \quad S = H P_{t_k}^l H^\transpose + \bar{N}_k .
    \end{equation}
\end{enumerate}
\end{mdframed}
Given these equations, once we know $A_t^l$ and $H$ matrices, we can implement the LI-EKF.

\subsection{Right-Invariant Extended Kalman Filter}
\label{sec:ri-ekf}
Next, we derive the Right-Invariant Extended Kalman Filter (RI-EKF) similarly. 

\subsubsection{Propagation}
In the propagation step, the mean can be directly computed using the process model. However, for the covariance propagation, we use the right-invariant log-linear error dynamics. As such, the state's mean evolves on the group while the covariance is tracked in the Lie algebra. 
    \begin{align}
        \nonumber \frac{d}{dt} \bar{X}_t &= f_{u_t}(\bar{X}_t), \quad t_{k-1} \leq t < t_k, \\
    % \end{equation*}
    % \begin{equation*} 
        \nonumber \frac{d}{dt} \eta_t^r &= g_{u_t}(\eta_t^r) - (\mathrm{Ad}_{\bar{X}_t} w_t)^\wedge \eta_t^r ,\\
        \nonumber &\implies \frac{d}{dt} \xi_t^r = A_t^r \xi_t^r - \mathrm{Ad}_{\bar{X}_t} w_t, \\
    % \end{equation*}
    % \begin{equation*} 
        \frac{d}{dt} P_t^r &= A_t^r P_t^r + P_t^r {A_t^r}^\transpose + \mathrm{Ad}_{\bar{X}_t} Q_t \mathrm{Ad}_{\bar{X}_t}^\transpose .
    \end{align}

\subsubsection{Update}
We use \mbox{$\eta = \exp(\xi) \approx I + \xi^\wedge$} and neglect the higher order terms to derive the linear update error equation as follows.
    \begin{align} 
        \nonumber \bar{X}_{t_k}^+ &= \exp \left( L_{t_k} \left( \bar{X}_{t_k} Y_{t_k} - b \right) \right) \bar{X}_{t_k}, \\
    % \end{equation*}
    % \begin{equation*} 
        \nonumber \eta_{t_k}^{r+} &= \exp \left( L_{t_k} \left( \eta_{t_k}^{r} b - b + \bar{X}_{t_k} V_t \right) \right) \eta_{t_k}^{r} , \\
    % \end{equation*}
    % \begin{equation*} 
    %     I + {\xi_{t_k}^{r+}}^\wedge = \left(I + \left( L_{t_k} \left( (I + {\xi_{t_k}^{r}}^\wedge) b - b + \bar{X}_{t_k} V_t \right) \right)^\wedge \right) (I + {\xi_{t_k}^{r}}^\wedge)
    % \end{equation*}
    % \begin{equation*} 
           \nonumber  {\xi_{t_k}^{r+}}^\wedge &= {\xi_{t_k}^{r}}^\wedge + \left( L_{t_k} \left( (I + {\xi_{t_k}^{r}}^\wedge) b - b + \bar{X}_{t_k} V_t \right) \right)^\wedge , \\
    % \end{equation*}
    % \begin{equation*} 
            \xi_{t_k}^{r+} &= \xi_{t_k}^{r} + L_{t_k} \left( {\xi_{t_k}^{r}}^\wedge b + \bar{X}_{t_k} V_t \right).
    \end{align}

To arrange all terms according to the vector form of $\xi$, define the measurement Jacobian, $H$, such that 
$\boxed{H \xi = -\xi^\wedge b}.$
Then it follows that
    \begin{align} 
        \nonumber  \bar{X}_{t_k}^+ &= \exp \left( L_{t_k} \left( \bar{X}_{t_k} Y_{t_k} - b \right) \right) \bar{X}_{t_k} , \\
    % \end{equation*}
    % \begin{equation*} 
    %         \xi_{t_k}^{r+} = \xi_{t_k}^{r} - L_{t_k} H \xi_{t_k}^{r} + L_{t_k} \bar{X}_{t_k} V_t
    % \end{equation*}
    % \begin{equation*} 
            \nonumber  \xi_{t_k}^{r+} &= (I - L_{t_k} H ) \xi_{t_k}^{r} + L_{t_k} \bar{X}_{t_k} V_t , \\
    % \end{equation*}
    % \begin{equation*}
        P_{t_k}^{r+} &= (I - L_{t_k} H) P_{t_k}^r (I - L_{t_k} H)^\transpose + L_{t_k} \bar{N}_k L_{t_k}^\transpose,
    \end{align}
    where
    % \begin{equation*}
        $\bar{N}_k := \bar{X}_{t_k} \operatorname{Cov}(V_k) \bar{X}_{t_k}^\transpose$.
    % \end{equation*}

\subsubsection{RI-EKF Result}
To summarize, we have the following two steps. 
\begin{mdframed}
\begin{enumerate}
  \item RI-EKF Propagation:
    \begin{align} 
        \label{eq:ri-prop-mean-cov}
         \nonumber  \frac{d}{dt} \bar{X}_t &= f_{u_t}(\bar{X}_t), \quad t_{k-1} \leq t < t_k, \\
    % \end{equation}
    % \begin{equation} 
    %     \label{eq:ri-prop-cov}
        \frac{d}{dt} P_t^r &= A_t^r P_t^r + P_t^r {A_t^r}^\transpose + \mathrm{Ad}_{\bar{X}_t} Q_t \mathrm{Ad}_{\bar{X}_t}^\transpose .
    \end{align}
  \item RI-EKF Update: 
    \begin{align} 
    \label{eq:ri-update-mean-cov}
         \nonumber  \bar{X}_{t_k}^+ &= \exp \left( L_{t_k} \left( \bar{X}_{t_k} Y_{t_k} - b \right) \right) \bar{X}_{t_k}, \\
    % \end{equation}
    % \begin{equation}
    % \label{eq:ri-update-cov}
        P_{t_k}^{r+} &= (I - L_{t_k} H) P_{t_k}^r (I - L_{t_k} H)^\transpose + L_{t_k} \bar{N}_k L_{t_k}^\transpose,
    \end{align}
    where 
    \begin{equation}
        L_{t_k} = P_{t_k}^r H^\transpose S^{-1}, \quad S = H P_{t_k}^r H^\transpose + \bar{N}_k .
    \end{equation}
 \end{enumerate}
\end{mdframed}
Given these equations, once we know $A_t^r$ and $H$ matrices, we can implement the RI-EKF.

\subsection{Switching Between Left and Right-Invariant Errors}
We can switch between the left and right error forms through the use of the adjoint map.
\begin{align}
% \begin{split}
    \nonumber \eta_t^r &= \bar{X}_t X_t^{-1} = \bar{X}_t \eta_t^l \bar{X}_t^{-1}, \\
    \nonumber \exp(\xi_t^r) &= \bar{X}_t \exp(\xi_t^l) \bar{X}_t^{-1} = \exp(\mathrm{Ad}_{\bar{X}_t} \xi_t^l), \\
    \xi_t^r &= \mathrm{Ad}_{\bar{X}_t} \xi_t^l .
% \end{split}
\end{align}

\begin{remark}
    This transformation is exact, which means that we can easily switch between the covariance of the left and right invariant errors using
    % \begin{equation*} 
        $P_t^r = \mathrm{Ad}_{\bar{X}_t} \, P_t^l \, \mathrm{Ad}_{\bar{X}_t}^\transpose$.
% \end{equation*}
\end{remark}

\begin{figure}[t]
\centering 
\includegraphics[trim={4cm 3.5cm 5cm 2.25cm},clip,width=1.00\columnwidth]{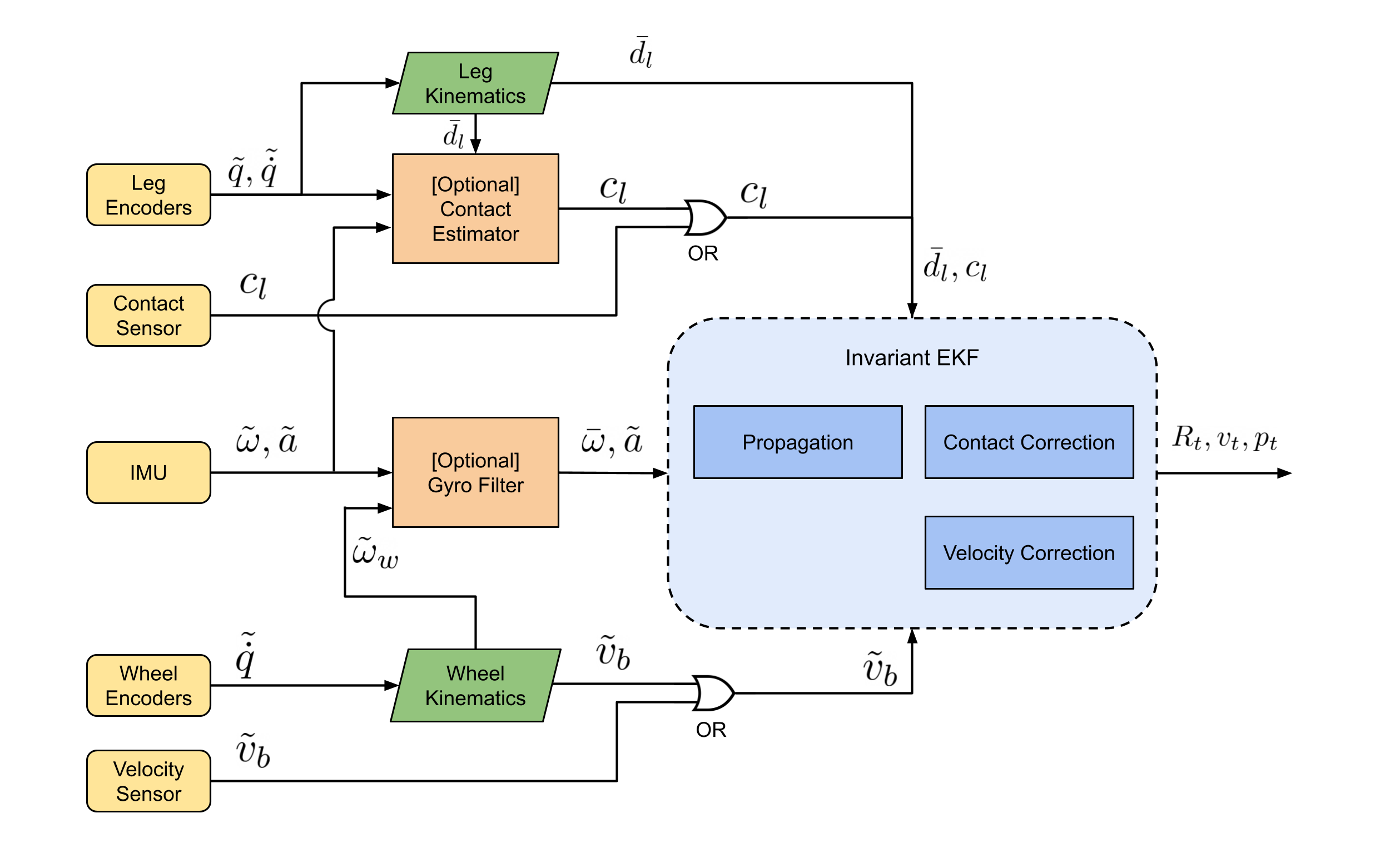}
\caption{%A diagram of the system structure. 
DRIFT takes measurements from an IMU and encoders as inputs. The angular velocities and linear accelerations are used in the propagation model. The encoder measurements are passed through kinematic functions and applied in the correction model. Two optional modules, a contact estimator and a gyro filter, are provided for low-cost robots to enhance their performance. }
\label{fig:drift}
\squeezeup
\end{figure}

\section{DRIFT: A Symmetry-Preserving Robot State Estimation Library}
\label{sec:drift}
% \mgj{add an into line and refer to drift Fig.~\ref{fig:drift}. Can't have floats not mentioned in the text.}
DRIFT can be used as a standalone C++ program, and we additionally provide an optional Robot Operating System (ROS) wrapper for easy communication between programs. The library can also be easily expanded with new sensor modalities. In addition, we provide two additional modules, a contact estimator and a gyro filter, to enable the proposed framework to be applied to low-cost robotic platforms. Fig.~\ref{fig:drift} shows the architecture implemented in this work.
% It is worth noticing that although the current implementation is based on the state-of-the-art InEKF, 
The proposed framework can be seamlessly extended to other existing and emerging symmetry-preserving filters in the future. 

\subsection{State Representation}
In most robotics applications, we are interested in tracking the orientation, velocity, and position of the robot with respect to a fixed world frame. We denote $R_t \in \mathrm{SO}(3)$, the rotation matrix that takes the coordinate frame from the body to the world, $v_t \in \mathbb{R}^3$, and $p_t \in \mathbb{R}^3$ as the body velocity and position in the world frame. Following~\citet{barrau2015non}, we can define our state as a direct isometries group $X_t \in \mathrm{SE}_{l+2}(3)$:
\begin{equation}
\label{eq:state}
    X_t \coloneqq \begin{bmatrix}
        R_t & v_t & p_t & d_{1t} & \cdots & d_{lt}\\
        0_{l+2,3} & &  & I_{l+2} %\\
        % 0 & 0 & 1 & 0 & \cdots & 0\\
        % 0 & 0 & 0 & 1 & \cdots & 0\\
        % \vdots & \vdots & \vdots & \vdots & \ddots & \vdots\\
        % 0 & 0 & 0 & 0 & \cdots & 1
    \end{bmatrix},
\end{equation}
% \begin{equation}
% \label{eq:state}
%     X_t \coloneqq \begin{bmatrix}
%         R_t & v_t & p_t & d_{1t} & \cdots & d_{lt}\\
%         0 & 1 & 0 & 0 &\cdots & 0\\
%         0 & 0 & 1 & 0 & \cdots & 0\\
%         0 & 0 & 0 & 1 & \cdots & 0\\
%         \vdots & \vdots & \vdots & \vdots & \ddots & \vdots\\
%         0 & 0 & 0 & 0 & \cdots & 1
%     \end{bmatrix},
% \end{equation}
where $d_{lt} \in \mathbb{R}^3$ is the augmented vectors, e.g., foot contact position in the world frame for the legged robot or landmark positions. In the minimal setup for wheeled or marine robots, the navigation state simply reduces to $\mathrm{SE}_2(3)$.% as in Example~\ref{example_imugps}.

\subsection{IMU Propagation}
% Thanks to the rapid development of MEMs sensors in the 2000s, IMUs are now cheap and becoming one of the default sensors on most robotic platforms. 
An IMU measures the angular velocity and the linear acceleration in the body frame. By integrating the measurements from an IMU, we can predict the robot's motion, enabling filter propagation without dealing with complicated dynamics~\citep{roumeliotis1999circumventing,hartley2020contact}.

In DRIFT, we model the IMU as corrupted by additive white Gaussian noise. 
\begin{align}
\begin{split}
    \label{eq:imu_measurement_unbias}
    \Tilde{\omega}_t &= \omega_t + w^g_t, \quad w^g_t \sim \mathcal{GP}(0_{3,1},\Sigma^g\delta(t-t')),\\
    \Tilde{a}_t &= a_t + w^a_t, \quad w^a_t \sim \mathcal{GP}(0_{3,1},\Sigma^a\delta(t-t')),
\end{split}
\end{align}
where, $\mathcal{GP}$ represents a Gaussian process and $\delta(t-t')$ is the Dirac delta function. With this, we define the input $u_t$ to the system as: 
% \begin{equation}
$u_t := \begin{bmatrix}
    \Tilde{\omega}_t^\transpose,
    \Tilde{a}_t^\transpose   
\end{bmatrix}^\transpose$.
% \end{equation}
\subsubsection{Continuous Dynamics} 
% Given the IMU measurements as inputs, 
We can write down the continuous dynamic process model:
\begin{equation}
    \frac{d}{dt}R_t = R_t(\Tilde{\omega}_t-w^g_t)_\times, 
    \frac{d}{dt}v_t = R_t(\Tilde{a}_t-w^a_t)+g , 
    \frac{d}{dt}p_t = v_t.
\end{equation}
Here, $(\cdot)_\times$ denotes a $3\times3$ skew-symmetric matrix, and $g$ is the known gravity vector.

For legged robots, 
% we might have additional contact states augmented. 
we assume the foot contact position in the world frame remains constant during the same contact period. As a result, the dynamics is only affected by the rotated white Gaussian noise:
\begin{equation}
    \label{eq:contact_propagation}
    \frac{d}{dt}d_t = R_t h_R(\Tilde{q}_t)(-w^d_t),
\end{equation}
where $\Tilde{q}_t$ is the encoder measurements, and $h_R(\Tilde{q}_t)$ is the orientation of the contact frame in the IMU (body) frame calculated from forward kinematics. For the sake of readability, we derive the propagation model with one contact augmentation in the following derivation. The propagation model for robots without foot contacts can be easily deduced by removing the corresponding column and row for the contact position.

Collecting the above equations into the matrix form and separating the noise term, we can obtain the deterministic dynamic function $f_{u_t}(\cdot)$:
\begin{align}
    \nonumber\frac{d}{dt} {X_t} &= 
    \begin{bmatrix}
        R_t(\Tilde{\omega}_t)_\times & R_t\Tilde{a}_t+g & v_t & 0 \\
        0 & 0 & 0 & 0 \\
        0 & 0 & 0 & 0 \\
        0 & 0 & 0 & 0 
    \end{bmatrix}\\
    \nonumber&-
    \begin{bmatrix}
        R_t & v_t & p_t & d_{t} \\
        0 & 1 & 0 & 0 \\
        0 & 0 & 1 & 0 \\
        0 & 0 & 0 & 1 
    \end{bmatrix}
    \begin{bmatrix}
        (w^g_t)_\times & w^a_t & 0 & h_R(\Tilde{q}_t)(w^v_t)\\
        0 & 0 & 0 & 0 \\
        0 & 0 & 0 & 0 \\
        0 & 0 & 0 & 0 
    \end{bmatrix}\\
    &\coloneqq f_{u_t}(X_t) - X_t w_t^\wedge .
\end{align}
We can verify that $f_{u_t}(\cdot)$ satisfies \eqref{eq:group_affine} and thus is group-affine, and from Theorem~\ref{theorem:autonomous_error_dynamics}, the right-invariant error dynamics is given by~\eqref{eq:error_dynamics}. % we obtain the autonomous error dynamics.
% \begin{equation}
%     \nonumber \frac{d}{dt} \eta_t^r = g_{u_t}(\eta_t^r) = f_{u_t}(\eta^r) - \eta^r f_{u_t}(I) .
% \end{equation}
% \tyl{I omit the derivation here. Do we want to include the derivation of A here?} 
Using $\nonumber g_{u_t}(\exp(\xi)) := (A_t \xi)^\wedge + \mathcal{O}(\lVert \xi \rVert^2)$ from Theorem~\ref{theorem:autonomous_error_dynamics} and following the derivation from Sec.~\ref{sec:ri-ekf}, we can obtain the state-independent linearized error dynamics:
\begin{align}
    \frac{d}{dt} \xi_t^r &= A_t^r \xi_t^r - \mathrm{Ad}_{\bar{X}_t} w_t, 
    A_t^r = \begin{bmatrix}
        0 & 0 & 0 & 0 \\
        (g)_\times & 0 & 0 & 0 \\
        0 & I & 0 & 0 \\
        0 & 0 & 0 & 0 \\
    \end{bmatrix}.
\end{align}
In summary, with $f_{u_t}(\cdot)$, $A_t^r$, and $Q_t = \operatorname{Cov}(w_t)$, the IMU propagation can then be computed using~\eqref{eq:ri-prop-mean-cov}.% and~\ref{eq:ri-prop-cov}:
% \begin{align*} 
%         \nonumber \frac{d}{dt} \bar{X}_t &= f_{u_t}(\bar{X}_t), \quad t_{k-1} \leq t < t_k, \\
%     % \end{equation}
%     % \begin{equation} 
%         \nonumber \frac{d}{dt} P_t^r &= A_t^r P_t^r + P_t^r {A_t^r}^\transpose + \mathrm{Ad}_{\bar{X}_t} Q_t \mathrm{Ad}_{\bar{X}_t}^\transpose.
%     \end{align*}
    
\subsubsection{Integration}
In practice, IMU provides discrete samplings of the continuous inputs. We assume the inputs to be constant between two timestamps and perform the propagation by integrating between time $t_k$ and $t_{k+1}$:
\begin{align}
    \nonumber \bar{R}_{t_{k+1}} &= \bar{R}_{t_k} \exp{({\omega}_{t_k}\Delta t)}\\
    \nonumber \bar{v}_{t_{k+1}} &= \bar{v}_{t_k} + g \Delta t + \bar{R}_{t_k}(\int_{t_k}^{t_{k+1}} \exp{(\omega_{t_k} t)} dt) {a}_{t_k}\\
    \nonumber \bar{p}_{t_{k+1}} &= \bar{p}_{t_k} + \bar{v}_{t_k} \Delta t + \frac{1}{2}g \Delta t^2\\
    &\quad + \bar{R}_{t_k}(\int_{t_k}^{t_{k+1}} \int_{t_k}^{t_\tau} \exp{(\omega_{t_k} t)} dt d\tau) {a}_{t_k} 
\end{align}
The exponential map and the integration for $\mathrm{SO}(3)$ has closed-form solutions~\cite{hartley2020contact,sola2017quaternion}, leading to
\begin{align}
    \nonumber \bar{R}_{t_{k+1}} &= \bar{R}_{t_k} \Gamma_0({\omega}_{t_k}\Delta t)\\
    \nonumber \bar{v}_{t_{k+1}} &= \bar{v}_{t_k} + g \Delta t + \bar{R}_{t_k}\Gamma_1({\omega}_{t_k}\Delta t) {a}_{t_k}\\
    \bar{p}_{t_{k+1}} &= \bar{p}_{t_k} + \bar{v}_{t_k} \Delta t + \frac{1}{2}g \Delta t^2 + \bar{R}_{t_k}\Gamma_2({\omega}_{t_k}\Delta t) {a}_{t_k},
\end{align}
with
\begin{align}
    \nonumber \Gamma_0(\phi) &= I + \frac{\sin(\| \phi \|)}{\| \phi \|}(\phi)_\times + \frac{1-\cos(\| \phi \|)}{\| \phi \|^2}(\phi)^2_\times\\
    \nonumber \Gamma_1(\phi) &= I + \frac{1-\cos(\| \phi \|)}{\| \phi \|^2}(\phi)_\times + \frac{\| \phi \|-\sin(\| \phi \|)}{\| \phi \|^3}(\phi)^2_\times\\
    \nonumber \Gamma_2(\phi) &= \frac{1}{2} I + \frac{\| \phi \|-\sin(\| \phi \|)}{\| \phi \|^3}(\phi)_\times\\
    & \quad + \frac{\| \phi \|^2 + 2\cos(\| \phi \|)-2}{2\| \phi \|^4}(\| \phi \|)^2_\times.
\end{align}
We can propagate the covariance using a discretized state transition matrix $\Phi^r = \exp(A^r \Delta t)$ and
\begin{align}
% \Phi^r &= \exp(A^r \Delta t)\\
 P_{k+1} = \Phi^r P_k \Phi^{r\transpose} + \mathrm{Ad}_{\bar{X}_k} Q_d \mathrm{Ad}_{\bar{X}_k}^\transpose .
\end{align}

\subsubsection{Bias Augmentation} 
In practice, IMU is often corrupted by slow-varying biases. To compensate for this, augmenting the biases as additional states in the system is common. However, no existing matrix Lie group can describe this additional state with the pose without violating the group-affine property~\citep{barrau2015non}. Adding IMU biases breaks the symmetry and makes the linearization matrix dependent on the estimated states~\citep{barrau2018invariant,hartley2020contact}. Nevertheless, this imperfect InEKF still outperforms traditional approaches like EKF~\citep{hartley2020contact}. 

To account for the biases, we model the IMU as being corrupted by additive low-frequency signals and white Gaussian noise:
\begin{align}
    \nonumber \Tilde{\omega}_t &= \omega_t + b^g_t + w^g_t, \quad w^g_t \sim \mathcal{GP}(0_{3,1},\Sigma^g\delta(t-t')),\\
    \Tilde{a}_t &= a_t + b^a_t + w^a_t, \quad w^a_t \sim \mathcal{GP}(0_{3,1},\Sigma^a\delta(t-t')).
\end{align}
The system's state then becomes a tuple of the state defined in~\eqref{eq:state} and the augmented biases:
\begin{align}
    (X_t, \theta_t)  \in \mathrm{SE}_{l+2}(3) \times \mathbb{R}^6, \ 
    \theta_t \coloneqq \begin{bmatrix}
        b^g_t\\
        b^a_t
    \end{bmatrix}.
\end{align}
The new augmented state has a right-invariant error in the form of:
% \begin{align}
    $e^r_t = (\bar{X_t}X_t^{-1},\bar{\theta}_t-\theta_t) \coloneqq (\eta_t,\zeta_t)$.
% \end{align}
With the bias augmented, we can obtain the continuous system dynamics:
\begin{align}
    \nonumber \frac{d}{dt}R_t &= R_t(\Tilde{\omega}_t-b^g_t-w^g_t)_\times, \\ 
    \nonumber \frac{d}{dt}v_t &= R_t(\Tilde{a}_t-b^a_t-w^a_t)+g , \\
    \frac{d}{dt}p_t &= v_t .
\end{align}
The contact position is not affected by biases. Therefore, for legged robots, we have the same contact dynamics as~\eqref{eq:contact_propagation}.
% \begin{align}
%     \frac{d}{dt}d_t &= R_t h_R(\Tilde{q}_t)(-w^d_t).
% \end{align}
Since the biases are slow-varying signals, we can model their dynamics using random walks. That is, white Gaussian noises govern the dynamics as follows.
\begin{align}
    \nonumber \frac{d}{dt}b^g_t &= w^{bg}_t, \quad w^{bg}_t \sim \mathcal{GP}(0_{3,1},\Sigma^{bg}\delta(t-t')),\\
    \frac{d}{dt}b^a_t &= w^{ag}_t, \quad w^{ag}_t \sim \mathcal{GP}(0_{3,1},\Sigma^{ag}\delta(t-t')).
\end{align}

With the new continuous dynamics, we can obtain our linearized error dynamics following Section \ref{sec:ri-ekf}:
\begin{align}
    \frac{d}{dt}\begin{bmatrix}
        \xi_t\\
        \zeta_t
    \end{bmatrix} &= A_t \begin{bmatrix}
        \xi_t\\
        \zeta_t
    \end{bmatrix} + \begin{bmatrix}
        Ad_{\bar{X}_t} & 0\\
        0 & I
    \end{bmatrix} w_t ,\\
    A_t & = \begin{bmatrix}
        0 & 0 & 0 & 0 & -\bar{R}_t & 0\\
        (g)_\times & 0 & 0 & 0 &  -(\bar{v}_t)_\times\bar{R}_t & -\bar{R}_t \\
        0 & I & 0 & 0  &  -(\bar{p}_t)_\times\bar{R}_t & 0\\
        0 & 0 & 0 & 0  &  -(\bar{d}_t)_\times\bar{R}_t & 0\\
        0 & 0 & 0 & 0  &  0 & 0
    \end{bmatrix}.
\end{align}

\subsection{Velocity Correction}
\label{sec:velocity_correction}
Body velocity measurements provide a generic correction model that can work on any robotic platform. Specifically, the filer requires the ground-referenced body velocity~\cite{teng2021legged,potokar2021invariant}. For some robotic applications, ground-referenced body velocity can be directly measured from the sensors, such as DVL for marine robots. However, accurate body velocity measurement is not always readily available. For wheeled robots with no lateral movements, one practical solution is to use the encoder measurements along with the nonholonomic constraints (i.e., velocity constraints that cannot be integrated) to form pseudo measurements for correction: 
\begin{equation}
    \label{eq:pseudo-velocity}
    v_w = \begin{bmatrix}
        \frac{r(\dot{q}_r+\dot{q}_l)}{2} & 0 & 0 
    \end{bmatrix}^\transpose,
\end{equation}
where $\dot{q}_r$ and $\dot{q}_l$ are angular velocity readings from the wheel encoders and $r$ is the wheel radius. The nonholonomic constraints provide the pseudo measurements in the $y$ and $z$ components under the assumption that the robot does not drift side-way nor jump up vertically. 
\subsubsection{Correction Model}
In DRIFT, the velocity correction model is implemented independently of the velocity source. This allows the library to be generic and to be applied to any robot with any source of velocity measurements. Here, we again assume the velocity measurements to be corrupted by white Gaussian noise:
\begin{equation}
    \Tilde{v}_t = v_t + w^v_t \quad w^v_t \sim \mathcal{GP}(0_{3,1},\Sigma^v\delta(t-t')).
\end{equation}
The body velocity is measured in the body frame. As a result, a right-invariant measurement model is used here. 
\begin{align}
    \nonumber Y_{t_k} &= X_{t_k}^{-1} b + V_{t_k}, \\
    \begin{bmatrix}
        \Tilde{v}_{t_k}\\
        -1\\
        0
    \end{bmatrix}
    &= \begin{bmatrix}
        R_{t_k}^\transpose & -R_{t_k}^\transpose v_{t_k} & -R_{t_k}^\transpose p_{t_k} \\
        0 & 1 & 0 \\
        0 & 0 & 1
    \end{bmatrix}
    \begin{bmatrix}
        0\\
        -1\\
        0
    \end{bmatrix}
    +\begin{bmatrix}
        w^v_{t_k}\\
        0\\
        0
    \end{bmatrix}.
\end{align}
We proceed to find $H$ as follows.
\begin{align*}
    H \xi_k^r &= -{\xi_k^r}^\wedge b \\
    H \begin{bmatrix} \xi^\omega_k \\ \xi^{v}_k \\ \xi^{p}_k \end{bmatrix} &= -\begin{bmatrix} {\xi^\omega_k}^\wedge & \xi^{v}_k & \xi^{p}_k \\ 0 & 0 & 0 \\ 0 & 0 & 0 \end{bmatrix}  \begin{bmatrix} 0 \\ -1 \\ 0  \end{bmatrix} =  \begin{bmatrix} \xi^{v} \\ 0 \\ 0 \end{bmatrix}\\
    H &= \begin{bmatrix} 
    0_{1,3} & I & 0  \\
    0_{1,3} & 0 & 0 \\
    0_{1,3} & 0 & 0 \end{bmatrix}.
\end{align*}
Then, we can perform the update step using~\eqref{eq:ri-update-mean-cov}, % and~\eqref{eq:ri-update-cov}:
% \begin{equation*} 
%     \bar{X}_{t_k}^+ = \exp \left( L_{t_k} \left( \bar{X}_{t_k} Y_{t_k} - b \right) \right) \bar{X}_{t_k}
% \end{equation*}
% \begin{equation*}
%     P_{t_k}^{r+} = (I - L_{t_k} H) P_{t_k}^r (I - L_{t_k} H)^\transpose + L_{t_k} \bar{N}_k L_{t_k}^\transpose
% \end{equation*}
% \begin{equation*}
%     L_{t_k} = P_{t_k}^r H^\transpose S^{-1}, \quad S = H P_{t_k}^r H^\transpose + \bar{N}_k .
% \end{equation*}
where after eliminating zeros
% \begin{equation*}
    $H = \begin{bmatrix} 0_{1,3} & I & 0  \end{bmatrix}$, and $N = \bar{R}_{t_k} \operatorname{Cov}(w^v)\bar{R}_{t_k}^\transpose$. 
% \end{equation*}

\subsubsection{Observability Analysis}
Theorem~\ref{theorem:log_linear_error} reveals the log-linear property of the error dynamics. Consequently, one can perform linear observability analysis using the linear error dynamic matrix~\citep{barrau2015non}, which, in our case, is time-invariant and nilpotent (with a degree of three): 
\begin{align*}
    \Phi &= \exp(A \Delta t) = \begin{bmatrix}
             I & 0 & 0  \\
             (g)_\times \Delta t & I & 0  \\
             \frac{1}{2}(g)_\times \Delta t^2 & 0 & I  \\
         \end{bmatrix}.
\end{align*}
Accordingly, the observability matrix becomes:
\begin{equation*}
\mathcal{O}=\begin{bmatrix}
{H} \\
{H} {\Phi} \\
{H} {\Phi}^{2} \\
\vdots
\end{bmatrix} = \begin{bmatrix}
{0} & {I} & {0} \\
(g)_\times \Delta t & {I} & 0  \\
2(g)_\times \Delta t & {I}& 0  \\
\vdots & \vdots & \vdots 
\end{bmatrix}.
\end{equation*}
% From the observability matrix, we can see that 
The roll and pitch are observable, as well as the velocity. However, since gravity only contains a value in the $z$ axis, the yaw angle is not observable. 

\subsection{Contact Correction}
Foot contacts provide additional constraints for legged robot state estimation by assuming the contact foot velocity in the world frame to be zero over the contact period. This allows the estimator to observe the body velocity from the forward kinematics models. In fact, this has become a core module for many modern legged robot state estimation algorithms~\citep{bloesch2012state,hartley2020contact,camurri2020pronto}. 
In DRIFT, we integrate the contact-inertial process model for legged robots proposed by~\citet{hartley2020contact}. 
% We briefly review the contact correction method. For detailed derivations, we refer the reader to~\citep{hartley2020contact}.
We assume the encoder measurements to be affected by zero-mean white Gaussian noise:
\begin{equation}
     \Tilde{q}_t = q_t + w^q_t, \quad w^q_t \sim \mathcal{GP}(0_{3,1},\Sigma^q\delta(t-t')).
\end{equation}
\subsubsection{Contact State Augmentation}
% When a contact is first detected by the contact estimator, it is appended to the state variables using the forward kinematics and the current position estimate:
When a contact point is detected, we append it to the state using the forward kinematics and the current position estimate:
\begin{equation*}
    \bar{d}_{lt} = \bar{p_t} + \bar{R_t}h_p(\Tilde{q}_t),
\end{equation*}
where $h_p(\cdot)$ is the forward kinematics function that maps the encoder measurements to the foot position in the body frame. 
The corresponding covariance can be augmented using
\begin{align*}
    P^{\mathrm{new}}_{t_k} &= F_{t_k}P_{t_k}F_{t_k}^\transpose + G_{t_k} \operatorname{Cov}(w_{t_k}^q) G_{t_k}^\transpose,\\
    F_{t_k} &= \begin{bmatrix}
        I & 0 & 0\\
        0 & I & 0\\
        0 & 0 & I\\
        0 & 0 & I\\
    \end{bmatrix}, \quad G_{t_k} = \begin{bmatrix}
        0\\
        0\\
        0\\
        \bar{R}_{t_k} J_p(\Tilde{q}_{t_k}),
    \end{bmatrix}
\end{align*}
where $J_p(\cdot)$ is the Jacobian of $h_p(\cdot)$. 

The augmented contact state remains in the state throughout the contact phase. When the contact constraint breaks (i.e., when the foot is lifted), the contact state is marginalized via
\begin{align}
    \bar{X}^{\mathrm{new}}_{t_k} = M \bar{X}_{t_k}, 
    \bar{P}^{\mathrm{new}}_{t_k} = M \bar{P}_{t_k} M^\transpose, 
    M = \begin{bmatrix}
        I & 0 & 0 & 0\\
        0 & I & 0 & 0\\
        0 & 0 & I & 0\\
    \end{bmatrix}.
\end{align}

\subsubsection{Correction Model}
Once the foot enters the contact phase, the augmented contact position follows~\eqref{eq:contact_propagation} in the propagation step. When a new encoder measurement is obtained, the right-invariant correction model is given by
\begin{align*}
    Y_{t_k} &= X_{t_k}^{-1} b + V_{t_k}\\
    \begin{bmatrix}
        h_p(\Tilde{q}_{t_k})\\
        0\\
        1\\
        -1
    \end{bmatrix}
    &= \begin{bmatrix}
        R_{t_k}^\transpose & -R_{t_k}^\transpose v_{t_k} & -R_{t_k}^\transpose p_{t_k} & -R_{t_k}^\transpose d_{t_k} \\
        0 & 1 & 0 & 0 \\
        0 & 0 & 1 & 0 \\
        0 & 0 & 0 & 1
    \end{bmatrix}
    \begin{bmatrix}
        0\\
        0 \\
        1 \\
        -1
    \end{bmatrix}\\
    &+\begin{bmatrix}
        J_p(\Tilde{q}_{t_k})w^q_{t_k}\\
        0\\
        0
    \end{bmatrix},
\end{align*}
The linearized measurement matrix $H$ and the noise matrix $N$ are as follows.
\begin{align*}
    H &= \begin{bmatrix}
        0 & 0 & -I & I
    \end{bmatrix},\\
    N &= \bar{R}_{t_k} J_p(\Tilde{q}_{t_k}) \operatorname{Cov}(w^q_{t_k}) J_p(\Tilde{q}_{t_k})^\transpose \bar{R}_{t_k}^\transpose .
\end{align*}

\subsection{Contact Estimator}
The contact correction model provides a means to correct the predicted state reliably. However, obtaining reliable contact estimation is often challenging, and false contact detection can introduce additional biases into the system. Moreover, for some low-cost legged robots, foot contact sensors are often not readily available. In the spirit of making the proposed framework more generalizable to different robotic platforms, we propose an additional contact detection module for robots without dedicated contact sensors~\citep{pmlr-v164-lin22b}. 

The proposed contact estimator is a self-supervised lightweight neural network that only takes measurements from an IMU and joint encoders as input. The neural network can operate at $830 \Hz$ on an NVIDIA Jetson AGX Xavier, which can be easily attached to existing robots. 
\subsubsection{Contact State Representation}
To design the network, we define our contact state $C$ as a vector of binary values:
\begin{align*}
    c &= \begin{bmatrix} c_{RF} & c_{LF} & c_{RH} & c_{LH} \end{bmatrix}, \quad &\text{for quadruped robots}\\
    c &= \begin{bmatrix} c_{R} & c_{L}\end{bmatrix}, \quad &\text{for bipedal robots}
\end{align*} where $c_l \in \{0,1\}$ for each leg $l$, with $1$ indicate a firm contact and $0$ as no contact. 

Depending on the robot's motion, the contact states can have a finite set of combinations, from all feet on the ground to all in the air. As a result, we can address the problem using a classification framework. That is, for each timestamp, we estimate which foot configurations are posed by the robot. In order to follow typical classification pipelines, we map our contact state vector $c$ to an integer $S$ by treating $c$ as a binary value and using binary-to-decimal conversion as the function. For quadruped robot, $S \in \{0,1,\dots,15\}$, and for bipedal, $S \in \{0,1,2,3\}$.

\subsubsection{Input Data}
The contact estimator depends solely on proprioceptive measurements, including joint encoders, IMU, and kinematics. For a synchronized time $n$, we concatenated the above measurements as
% \begin{equation*}
    $z_n= 
    \begin{bmatrix}
        q_n & \dot{q}_n & a_n & \omega_n & d_{ln} & \dot{d}_{ln} 
    \end{bmatrix}$.
% \end{equation*}
In order for the network to incorporate temporal information, we append previous $w-1$ measurements to the current data: $D_n = 
    \begin{bmatrix}
        z_{n-w}^\transpose & z_{n-w+1}^\transpose & \hdots & z_{n}^\transpose 
    \end{bmatrix}^\transpose$. Therefore, at each time $n$, the network takes in a 2D matrix $D_n$ and estimates the current contact configuration $S_n$.

\subsubsection{Network Structure} 
% \tyl{This paragraph is not smooth. TODO: Rewrite it.}

The network comprises 2 convolutional blocks and 3 fully connected layers. Each convolutional block consists of 2 one-dimensional convolutions and a one-dimensional max pooling. We choose a one-dimensional kernel to improve efficiency and reduce memory usage. For nonlinearity, we employ ReLU as the activation function. To prevent the network from overfitting, we add a dropout mechanism at the end of the second convolutional block. 

With the deep features extracted from the convolutional blocks, the fully connected layers convert them into the contact configuration $S_n$. Again, to prevent the network from overfitting, we employ dropout mechanisms for the first 2 fully connected layers. Finally, the cross-entropy loss is applied for the classification task: 
\begin{equation}
    L(P_i) = -\log{\frac{\exp{(P_i)}}{\sum_j{\exp{(P_j)}}}},
\end{equation}
where $P_j$ is the probability output from the network of state $j$, and $P_i$ is the probability of the ground truth state.
% \mgj{explanation of the self-supervised labels is missing. Just take it from the corl paper.}

% \subsection{}

\subsection{Gyro Filter}
\label{sec:gyro_filter}
% Some robots are only equipped with 
Low-cost IMUs 
% . These IMUs 
can deteriorate performance. This is especially worse for states that are not observable, such as the yaw angle. To mitigate this issue, we propose an additional gyro filter that fuses angular velocity measurements from different sources, such as additional IMUs or kinematic models. The gyro filter is a linear Kalman filter, with its state defined as:
\begin{equation}
    x := \begin{bmatrix}
        \omega^\transpose & {b^{g}}^\transpose
    \end{bmatrix}^\transpose,
\end{equation}
where $\omega$ is a $3$-vector representing the angular velocity in the base IMU frame, and $b^{g}$ denotes the corresponding biases. 
\subsubsection{Propagation}
In the propagation step, we assume the measurement biases to be slowing-varying quantities, which remain the same between two timestamps. As a result, we can formulate an integration model by taking the differences between two consecutive measurements and canceling out the biases as:
\begin{align}
x_{k+1} = x_{k} + \begin{bmatrix}
    \tilde{\omega}^{\alpha}_{k+1} - \tilde{\omega}^{\alpha}_{k},\\
    0_{3 \times 1}
\end{bmatrix}, P_k = P_k + Q_k.
\end{align}
Here, $\tilde{\omega}^{\alpha}$ denotes the angular velocity measurements from one sensing source, $P_k$ is the filter covariance, and $Q_k$ is the process noise covariance. 
\subsubsection{Correction}
A second source of angular velocity measurements, $\tilde{\omega}^{\beta}$, can be used to correct the filter using a linear measurement model:
\begin{align*}
    \tilde{\omega}^{\beta} = H x.
\end{align*}
The corresponding $H$ matrix is $\begin{bmatrix}
    I_{3 \times 3} & I_{3 \times 3} 
\end{bmatrix}$ if the measurement is biased, and $\begin{bmatrix}
    I_{3 \times 3} & 0_{3 \times 3} 
\end{bmatrix}$ if the measurement is unbiased. With this model, the filter can be corrected following the standard EKF correction algorithm~\citep{barfoot_2017}.

\begin{figure*}[t]
    \centering
    \subfloat{
        \includegraphics[width=0.4\textwidth]{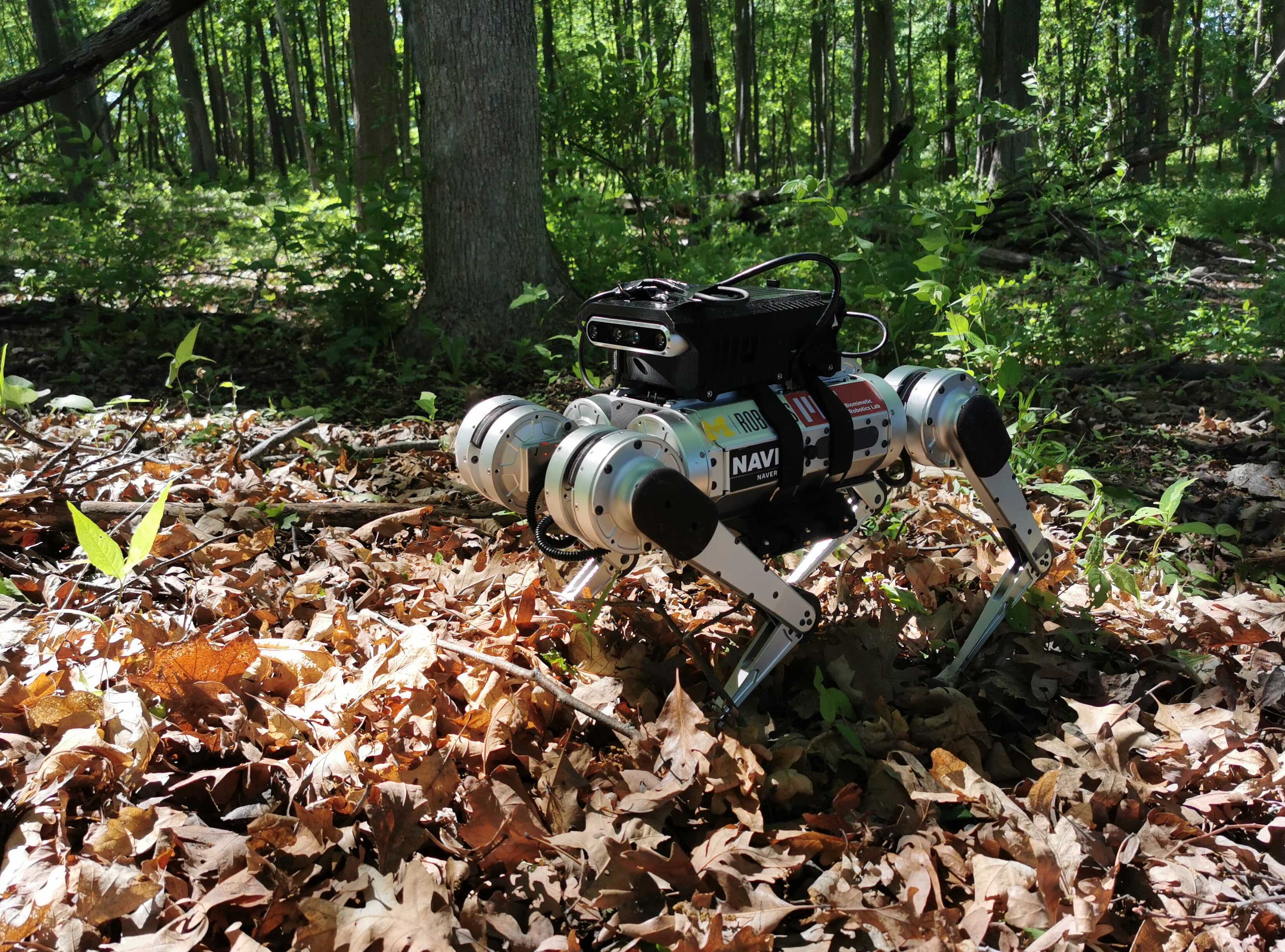}
        \label{fig:lab_setup}}
    \subfloat{
        \includegraphics[width=0.53\textwidth,trim={1cm 0.3cm 1cm 1cm},clip]{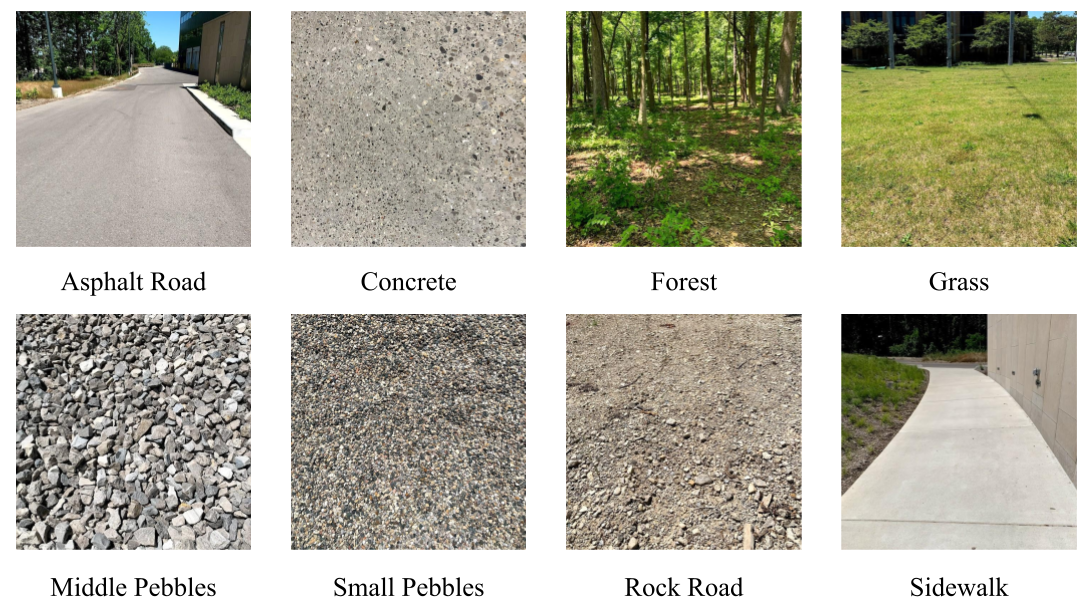}
        }
     \caption{Left: The configuration of an MIT Mini Cheetah robot for contact data collection. Right: Different terrain types in the contact data set.}
     \label{fig:terrain_type}
     \vspace{-4mm}
\end{figure*}

\section{Experimental Results}
\label{sec:experiments}
\subsection{Contact Estimation}
% \subsubsection{Mini Cheetah Contact Data Sets}
% In order 
To train the proposed contact estimator, we create open-source contact data sets using an MIT Mini Cheetah robot~\citep{katz2019mini}. We record all proprioceptive sensor measurements, including joint encoders data, foot position and velocities, IMU measurements, and estimated joint torques from an MIT controller~\citep{kim2019highly}. All data are upsampled to $1000 \Hz$ to match the IMU frequency. The data sets are created over 8 different terrains, including asphalt roads, concrete, forest, grass, middle pebbles, small pebbles, rock road, and sidewalks. In addition to the above terrains, several sequences of the robot walking in the air (i.e., not having contact with the ground while operating the same gait.) are recorded to provide negative examples to the network. We collect around 1,000,000 data points and reserve some sequences for state estimation tests. The rest of the data sets are separated into testing, validation, and training sets with the ratio of $15\%$, $15\%$, and $70\%$. Fig.~\ref{fig:terrain_type} shows examples of the different terrain types in the contact data sets. 

The ground truth labels are generated using an offline algorithm, which takes the robot's foot height in the hip frame as input. The algorithm applies a low-pass filter and uses future and past data points around the current time stamp to extract local minima and maxima. Moreover, we observe a bouncing effect on the robot's foot upon touchdown after inspecting slow-motion videos of Mini Cheetah's walking patterns. The bouncing results in a sudden change of foot height in the signal, causing false positives which can be removed by applying the low-pass filter. %to the signal enables the algorithm to remove these false positives from the ground truth.

% \subsubsection{Contact Estimation Results}
% We now 
We present the accuracy, false positive rate (FPR), and false negative rate (FNR) of the proposed method against other baselines. For baselines, we implement a model-based approach~\citep{focchi2013local,fakoorian2016ground,fink2020proprioceptive}, where the estimated ground reaction force (GRF) is computed via the general equation of motion with a low-pass filter. A fixed threshold is set to the estimated GRF to detect the contact events. We would like to highlight that because there is no direct access to the motor current on the Mini Cheetah robot, we use the torque command from the controller to approximate the actual torque on the actuators. We denote this method as GRF thresholding. In addition, we also obtain the gait cycle command from the MIT controller~\citep{kim2019highly} to serve as a second contact estimation baseline.

Table~\ref{tab:baseline_acc} lists the accuracy, FPR, and FNR of the compared methods on three test sequences. We can see that the proposed method achieves the highest accuracy across all sequences and the GRF thresholding has the worst performance. Although the baselines obtain slightly lower FNRs, our method maintains significantly lower FPRs. Lower FPR is crucial for state estimation tasks as false positive contact events can introduce biased measurements into the system.
\begin{table*}[t]
    \centering
    \caption{Accuracy comparison against baselines.}
    \footnotesize 
    \resizebox{1.99\columnwidth}{!}{
    \begin{tabular}{r r r r r r r c c}
        \toprule
        \textbf{Sequence} & \textbf{Method} & \multicolumn{5}{c}{\textbf{\% Accuracy}}  &
        \multicolumn{1}{c}{\textbf{\% False Positive Rate}} & \multicolumn{1}{c}{\textbf{\% False Negative Rate}} \\
        & &  Leg RF & Leg LF & Leg RH & Leg LH & Leg Avg & Leg Avg & Leg Avg \\
        \midrule
        \multirow{3}{*}{Concrete Test Sequence} & GRF Thresholding & 73.43 & 70.02 & 71.69 & 70.04 & 71.30  & 37.07 & 13.24\\
        & Gait Cycle & 85.66 & 84.98 &	84.68 &	85.11 &	85.11 &	22.95 & \textbf{0.00} \\
        & Proposed Method & \textbf{98.34} &	\textbf{97.87} &	\textbf{97.95} &	\textbf{98.56} &	\textbf{98.18} & \textbf{1.45} & 2.51 \\
        \midrule
        \multirow{3}{*}{Grass Test Sequence} & GRF Thresholding & 82.55 & 78.93 & 84.62 &	82.48 &	82.14 & 26.87 & \textbf{0.63} \\
        & Gait Cycle & 92.41 &	92.38 & 91.04 &	90.55 &	91.59 & 10.95 & 3.53 \\
        & Proposed Method & \textbf{98.08} &	\textbf{97.57} &	\textbf{97.73} &	\textbf{97.73} &	\textbf{97.78} & \textbf{2.35} & 1.98 \\
        \midrule
        \multirow{3}{*}{Forest Test Sequence} & GRF Thresholding & 80.99 &	80.09 &	82.75 & 83.24 &	81.77 & 26.54 & 1.84 \\
        & Gait Cycle & 83.03 &	82.56 &	84.44 &	84.28 &	83.58 & 24.71 & \textbf{0.08} \\
        & Proposed Method & \textbf{97.05} &	\textbf{96.62} &	\textbf{97.24} &	\textbf{97.40} &	\textbf{97.08} & \textbf{2.82} & 3.12 \\
        \bottomrule
    \end{tabular}}
    \label{tab:baseline_acc}
    \vspace{-4mm}
\end{table*}
\begin{figure*}[t]
\vspace{-2mm}
\centering
    \subfloat{
        \includegraphics[width=0.41\textwidth]{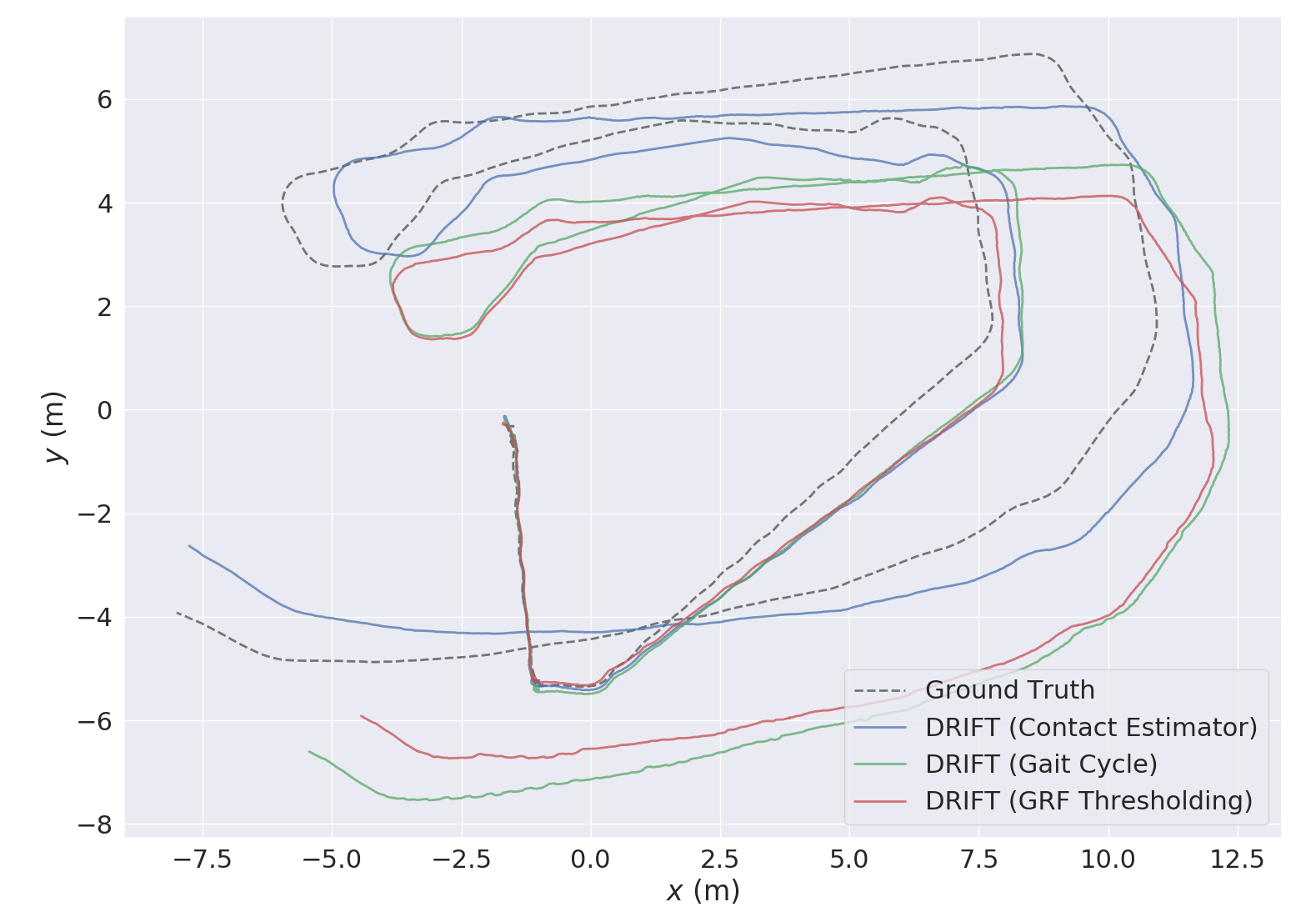}
        \label{fig:mini_cheetah_mair_birds_eye}}
    \subfloat{
    \includegraphics[width=0.55\textwidth]{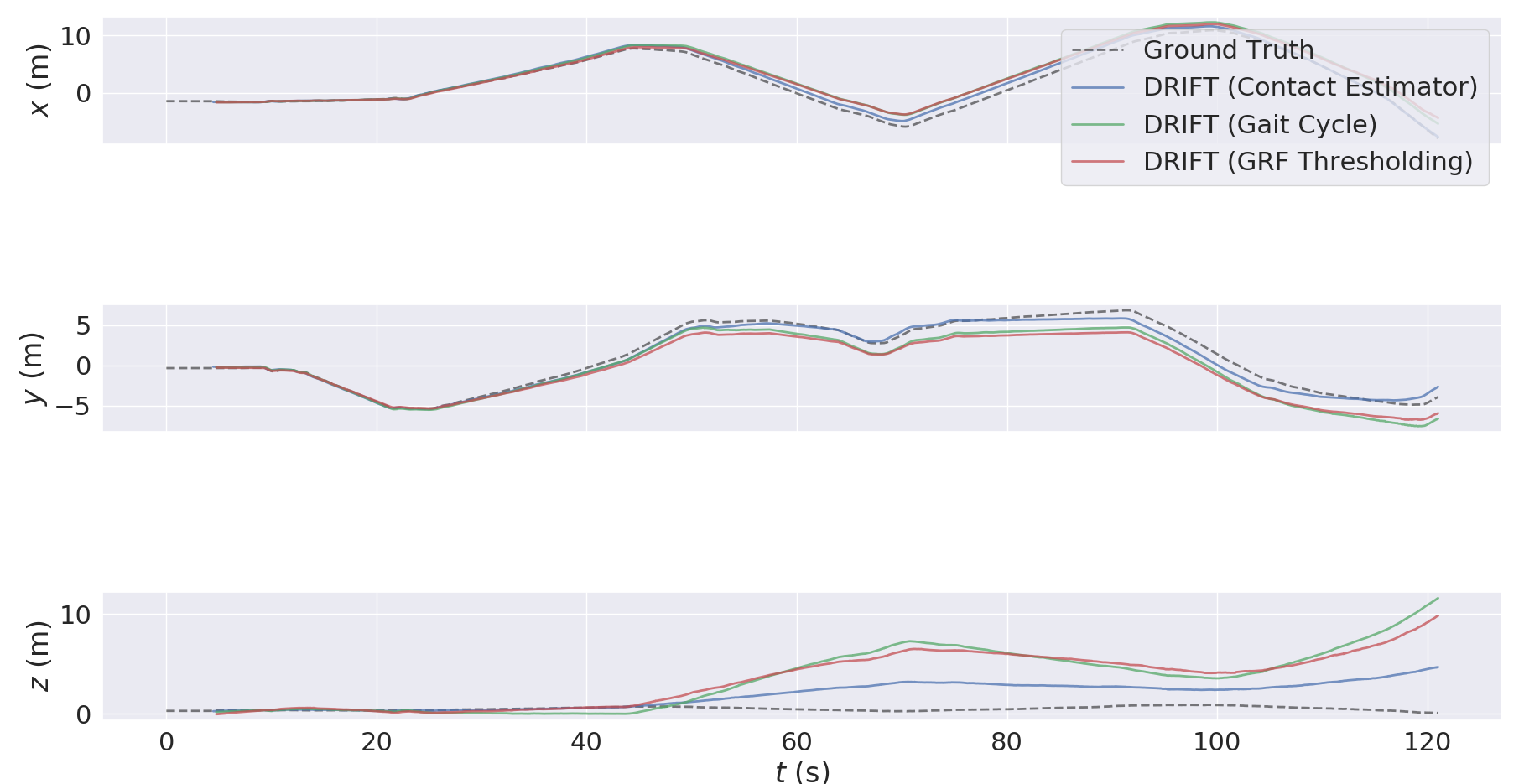}
        \label{fig:mini_cheetah_mair_xyz}}
     \caption{The estimated trajectories from DRIFT using different contact estimation methods on the Mini Cheetah data set. The robot walks on a grassy field with a motion capture system, which is used for ground truth capturing. With the proposed contact estimator, DRIFT produces the best trajectory. Contrarily, the two baseline methods introduce significant drift in the height ($z$) axis. This figure is generated with the aid of the Python package evo~\citep{grupp2017evo}.}
     \label{fig:mini_cheetah_mair}
\end{figure*}

\subsection{Legged Robot}
We present the state estimation results using DRIFT with the proposed contact estimator for the Mini Cheetah robot. The robot walks on an outdoor grassy field, where we use a motion capture system to obtain the ground truth pose. We deploy the proposed contact estimator, as well as the two baseline contact estimation methods with DRIFT. Fig.~\ref{fig:mini_cheetah_mair} shows the estimated trajectory. Qualitatively, DRIFT with the proposed contact estimator produces a trajectory closer to the ground truth. In addition, the two baseline methods produce extra height ($z$) drifts. We argue this is the result of the high false positive rates of the baseline methods. This experiment confirms DRIFT's support for accurate legged robot state estimation. Moreover, the contact estimation module allows robots without contact sensors, such as the MIT Mini Cheetah, to obtain reliable contact estimation results, which are essential to contact-aided state estimation algorithms.

% \begin{figure}[ht]
%     \includegraphics[width=0.98\columnwidth]{figures/mini_cheetah_mair.png}
%     \caption{}
%     \label{fig:mini_cheetah_mair}
% \end{figure}

\begin{figure}[t]
\centering
    \includegraphics[width=0.8\columnwidth]{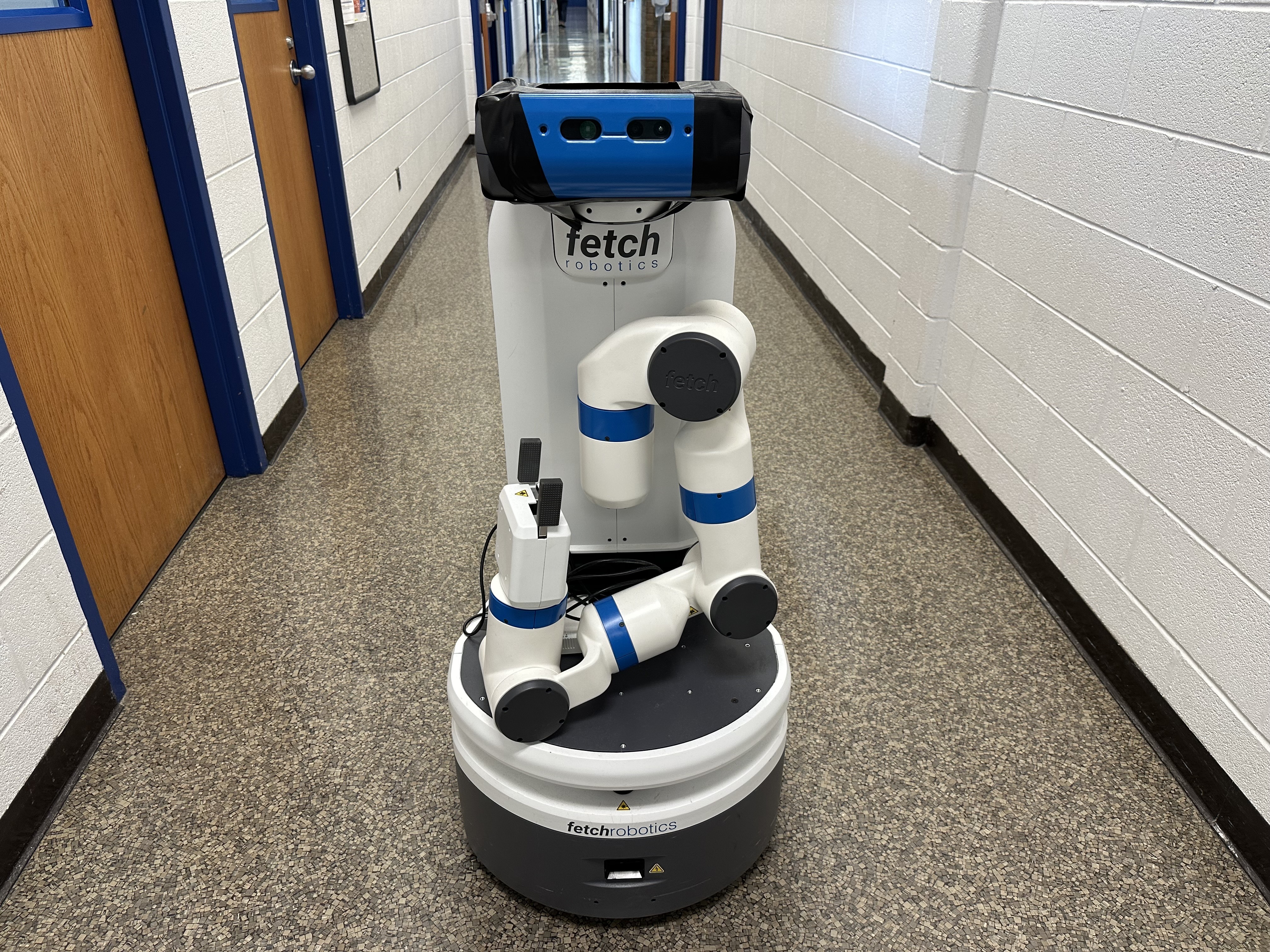}
    \caption{A Fetch robot used in the indoor experiments. The figure shows the robot at the Department of Naval Architecture and Marine Engineering 
 on the University of Michigan campus. Fetch is a differential drive robot that is commonly used for indoor service robot research.}
    \label{fig:fetch}
\end{figure}
\begin{table*}[t]
\caption{The RMSE of Relative Pose Error (RPE) for Fetch. We report the RPE in the unit of drift per meter.}
\begin{center}
\begin{tabular}{l l | c c c c c c c | c c}
    \toprule
     & &  01_square & 02_um & 03_eight & 04_random & 05_right_turn & 06_left_turn & 07_origin  & \textbf{Avg} & \textbf{Std}\\
     \midrule 
      & Trajectory & \multirow{2}{*}{10.59} & \multirow{2}{*}{14.15} & \multirow{2}{*}{11.61}& \multirow{2}{*}{38.91}& \multirow{2}{*}{15.64} & \multirow{2}{*}{22.29} & \multirow{2}{*}{11.26} &  \multirow{2}{*}{17.78} & \multirow{2}{*}{N/A}\\
      & Length (m) &&&&&&&\\
    \midrule
      & Duration ($\sec$) & 43.720 & 49.62 & 34.62 & 99.06 & 39.72 & 51.71 & 46.68 & 52.16 & N/A\\
    \midrule
    MEKF~\cite{sola2017quaternion}& Trans. ($\m/\m$) & 0.1012 & 0.0634 & 0.0855 & 0.0990 & 0.0853 & \textbf{0.0435} & 0.1126 & 0.0844 & 0.0182\\
          & Rot. ($\degree/\m$) & \textbf{1.0555} & 5.5752 & 3.8387 & 6.9809 & 3.9911 & 1.6304 & \textbf{2.4502} & 3.6460 & 1.1016\\
    \midrule
    DRIFT & Trans. ($\m/\m$) & 0.0688 & 0.0587 & 0.0629 & 0.0863 & 0.0857 & 0.0436 & 0.0786 & 0.0692 & 0.0156\\
          & Rot. ($\degree/\m$) & 1.9766 & \textbf{5.3704} & \textbf{3.6134} & \textbf{6.6857} & 3.6606 & \textbf{1.5580} & 2.4741 & 3.6198 & {1.8587}\\
    \midrule
    DRIFT & Trans. ($\m/\m$) & \textbf{0.0565} & \textbf{0.0585} & \textbf{0.0566} & \textbf{0.0801} & \textbf{0.0613} & 0.0438 & \textbf{0.0558} & \textbf{0.0590} & {0.0127}\\
    (Gyro Filter) & Rot. ($\degree/\m$) & 1.5723 & 5.3865 & 3.8561 & 6.7959 & \textbf{1.9193} & 1.9246 & 3.4866 & \textbf{3.5631} & 1.2172\\
    \bottomrule
\end{tabular}
\label{tab:fetch_rpe}
\end{center}
\vspace{-5mm}
\end{table*}

\subsection{Indoor Wheeled Robot}
For indoor applications, we deploy DRIFT on a Fetch robot.
% , which is commonly used for indoor service robot research. 
A picture of the Fetch robot is shown in Fig.~\ref{fig:fetch}. Fetch is equipped with a low-cost IMU and wheel encoders on each wheel, which we use as the input to DRIFT. We collected data sets in a lab environment, where Fetch drove on a polished concrete surface for several sequences. We obtained the ground truth poses from a motion capture system. We additionally implement a MEKF~\citep{sola2017quaternion} to serve as a baseline algorithm. Both DRIFT and MEKF take only IMU and encoder measurements as input and estimate the 6D pose of the robot. The low-cost IMU on Fetch can cause the unobservable yaw angle to degrade and further reduce the accuracy of the overall state estimation. As a result, we implemented an additional gyro filter described in Sec.~\ref{sec:gyro_filter} to fuse the yaw angular velocity measurement from the IMU with the encoder measurements. 

\begin{figure}[t]
\centering
    \includegraphics[width=0.8\columnwidth]{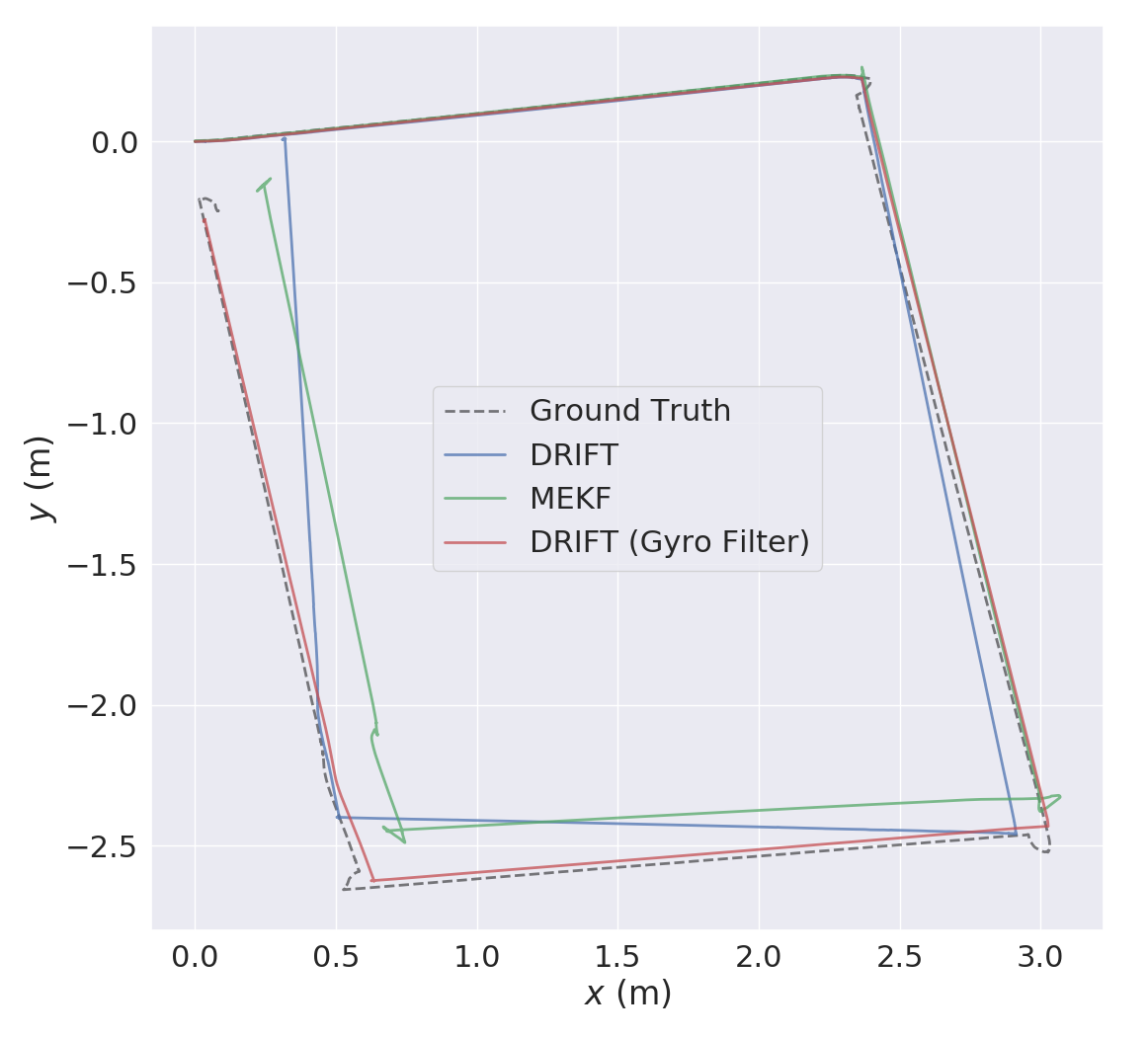}
    \caption{The bird's-eye view of the estimated trajectory on the \texttt{01_square} sequence in the Fetch data set. Overall, the estimated trajectory from DRIFT is smoother than MEKF. Among them, DRIFT with the gyro filter has the best performance.}
    \label{fig:fetch_qualitative}
\end{figure}

Fig.~\ref{fig:fetch_qualitative} shows a top-down view of the estimated trajectory on the \texttt{01_square} sequence. We observe that because of the inaccurate yaw angular velocity measurements, both DRIFT and MEKF experience significant drift after turning. However, with the help of the gyro filter, we can obtain a much better trajectory estimation. In addition, we notice that the estimated trajectory from DRIFT is much smoother than MEKF.

% We present the quantitative results on the Fetch data set. 
Since DRIFT is an odometry framework instead of a full SLAM system, we follow~\citet{sturm12iros} and report the root-mean-square error (RMSE) of the Relative Pose Error (RPE), in the unit of drift per meter in Table~\ref{tab:fetch_rpe}. DRIFT performs consistently better than the baseline. With the aid of the gyro filter, the position estimation is further improved, which likely results from a more accurate yaw angle estimation.

\begin{figure}[t]
\centering
    \includegraphics[width=0.75\columnwidth]{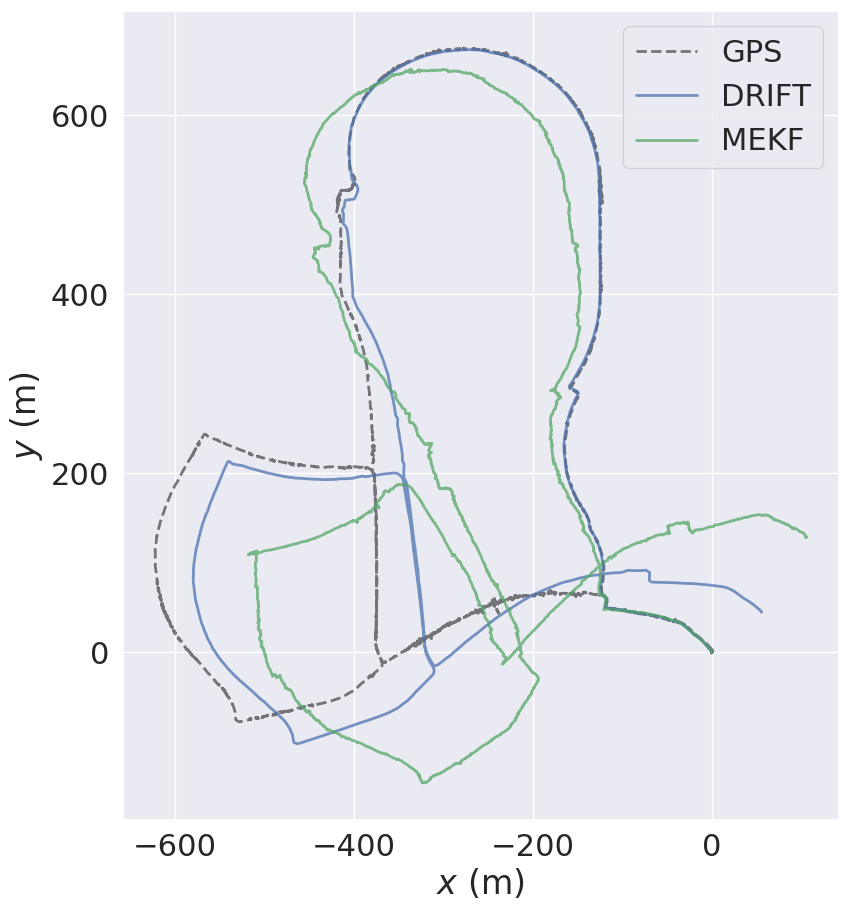}
    \caption{%The bird's eye view of the estimated trajectories in the 
    Husky long-horizon experiment. The robot starts and ends at the bottom right corner. While being purely proprioceptive, DRIFT can obtain accurate estimates after 3 kilometers of operation. 
    % This showcases that DRIFT can be a reliable odometry source for visual-SLAM systems in perceptually degraded environments, even for long-horizon operations.
    }
    \label{fig:husky_northwood_experiment}
    \squeezeup
\end{figure}

\begin{table*}[t]
\caption{The RMSE of Relative Pose Error (RPE) for Husky short-horizon experiment. We report the RPE in the unit of drift per meter.}
\begin{center}
\addtolength{\tabcolsep}{-2pt}
\begin{tabular}{l l | c c c c c c c c c | c c}
    \toprule
     & &  1_rectangle	& 2_batman & 3_broken_cat & 4_um & 5_type_z & 6_rocker & 7_peace & 8_cheetah & 9_m  & \textbf{Avg} & \textbf{Std}\\
     \midrule 
      & Trajectory & \multirow{2}{*}{20.56} & \multirow{2}{*}{38.47} & \multirow{2}{*}{90.43} & \multirow{2}{*}{39.11} & \multirow{2}{*}{59.89} & \multirow{2}{*}{39.46} & \multirow{2}{*}{40.56} & \multirow{2}{*}{50.37} & \multirow{2}{*}{63.72} & \multirow{2}{*}{49.17} & \multirow{2}{*}{N/A}\\
      & Length ($\m$) &&&&&&&&&\\
    \midrule
      & Duration ($\sec$) & 49.78 & 77.92 & 154.57 & 67.09 & 90.48 & 67.49 & 66.50 & 92.01 & 100.07 & 85.10 & N/A\\
    \midrule    
    MEKF~\cite{sola2017quaternion}& Trans. ($\m/\m$) & \textbf{0.0477} & 0.0812 & 0.1139 & 0.0625 & \textbf{0.0652} & 0.0680 & 0.0749 & 0.0829 & \textbf{0.0755} & 0.0747 & 0.0384\\
          & Rot. ($\degree/\m$) & 1.3738 & \textbf{4.8355} & 2.5033 & 1.6061 & 1.9539 & 1.6396 & 1.4865 & 1.6416 & 1.3962 & 2.0485 & {1.0650}\\
    \midrule
    DRIFT & Trans. ($\m/\m$) & 0.0482 & \textbf{0.0781} & \textbf{0.0881} & \textbf{0.0564} & 0.0678 & \textbf{0.0627} & \textbf{0.0728} & \textbf{0.0817} & \textbf{0.0755} & \textbf{0.0701} & {0.0351}\\
          & Rot. ($\degree/\m$) & \textbf{1.0152} & 4.8559 & \textbf{1.7097} & \textbf{1.3381} & \textbf{1.3089} & \textbf{1.3126} & \textbf{1.1606} & \textbf{1.6321} & \textbf{0.8657} & \textbf{1.6888} & 1.2657\\

    % \midrule
    % MEKF~\cite{sola2017quaternion} & Trans. ($\m/\m$) & 0.0777 & 0.1363 & 0.1971 & 0.1427 & 0.0881 & 0.1513 & 0.1255 & 0.1245 & 0.1062 & 0.1277 & 0.0384\\
    %    (Same param)   & Rot. ($\degree/\m$) & 1.5364 & 4.7861 & 3.8169 & 1.9029 & 2.2957 & 2.2517 & 2.0711 & 2.6845 & 1.8538 & 2.5777 & 1.0650\\
    \bottomrule
\end{tabular}
\label{tab:husky_mair_rpe}
\end{center}
\squeezeup\squeezeup
\end{table*}

\subsection{Outdoor Wheeled Robot}
% \begin{figure}[t]
% \centering
%     \includegraphics[width=0.9\columnwidth]{figures/husky.png}
%     \caption{A snapshot of the Husky robot in the long-horizon experiment. During this experiment, Husky was driven on the sidewalk for $3$ kilometers.}
%     \label{fig:husky}
% \end{figure}
% To evaluate outdoor scenarios, 
We deploy DRIFT on a Clearpath Robotics Husky robot. Husky is a four-wheeled differential-drive robot, as shown in Fig.~\ref{fig:northwood_satellite}. The robot is equipped with two VectorNav VN-100 IMUs and wheel encoders on each side. We conducted both short and long-horizon experiments using the Husky robot. The short-horizon data were collected on an outdoor grassy field with a motion capture system, which we used to obtain ground truth poses. For the long-horizon experiment, we drove the robot on the sidewalk for $55$ minutes. The total trajectory path is around 3 kilometers. Since no motion capture system is possible for such large-scale experiments, we used the GPS signals as a proxy for the ground truth trajectory. To better assess the performance of DRIFT, we again used a MEKF~\citep{sola2017quaternion} as the baseline algorithm. 

Table~\ref{tab:husky_mair_rpe} reports the RPE of the short-horizon experiment. On average, DRIFT performs better than the MEKF. This is expected as both the state-transition matrix $A$ and linearized measurement matrix $H$ of the InEKF are independent of the estimated states, while the corresponding matrices in the MEKF depend on the estimated states. 

The benefit of using InEKF becomes even more obvious in long-horizon operations. Fig.~\ref{fig:northwood_satellite} and~\ref{fig:husky_northwood_experiment} show the bird's-eye view of the estimated trajectory for the long-horizon experiment. We can see DRIFT performs significantly better than MEKF, with the final drift in the $xy$ plane to be $71.41 \m$ and $166.40 \m$, respectively. We highlight that DRIFT uses proprioceptive data only. This demonstrates DRIFT's capability of being a reliable odometry source for visual-SLAM systems in perceptually degraded environments, even for long-horizon operations. 

\begin{figure}[t]
\centering
    \includegraphics[width=0.99\columnwidth,trim={1.5cm 0.5cm 0.5cm 0.5cm},clip]{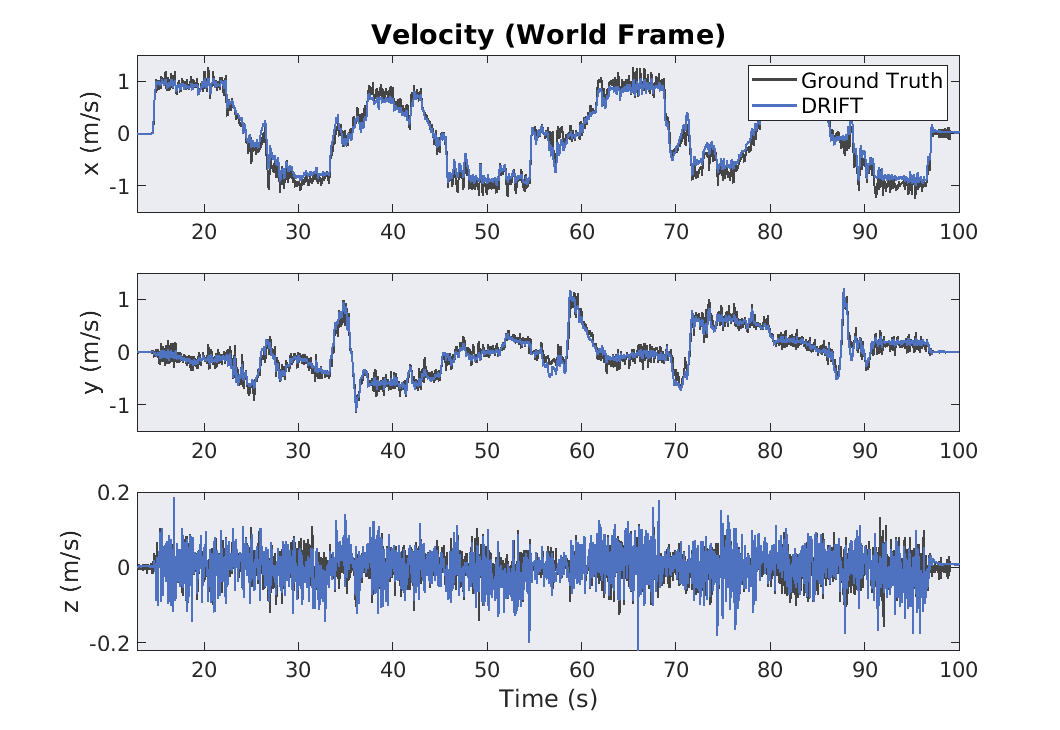}
    \caption{The velocity estimated from DRIFT overlapped with the ground truth velocity for the \texttt{9_M} sequence. The estimated velocity converges to the ground truth velocity, which agrees with the observability analysis in Sec.~\ref{sec:drift}.}
    \label{fig:husky_9_m_velocity}
    \squeezeup
\end{figure}

% In addition to evaluating the estimated pose, 
We also compare the estimated velocity in the world frame with the ground truth velocity in Fig.~\ref{fig:husky_9_m_velocity}. The ground truth velocity is obtained by differentiating the ground truth pose from the motion capture system. We see the estimated velocity converges to the true velocity, which agrees with the observability analysis in Sec.~\ref{sec:drift}.

% \begin{figure*}[ht]
%     \subfloat{
%         \includegraphics[width=0.47\textwidth]{figures/husky_9_m_velocity.png}
%         \label{fig:}}
%     \subfloat{
%         \includegraphics[width=0.51\textwidth]{figures/husky_9_m_rpy.png}
%         \label{fig:}}
%      \caption{}
%      \label{fig:husky_9_m}
% \end{figure*}

\begin{figure}[t]
\centering
    \includegraphics[width=0.99\columnwidth]{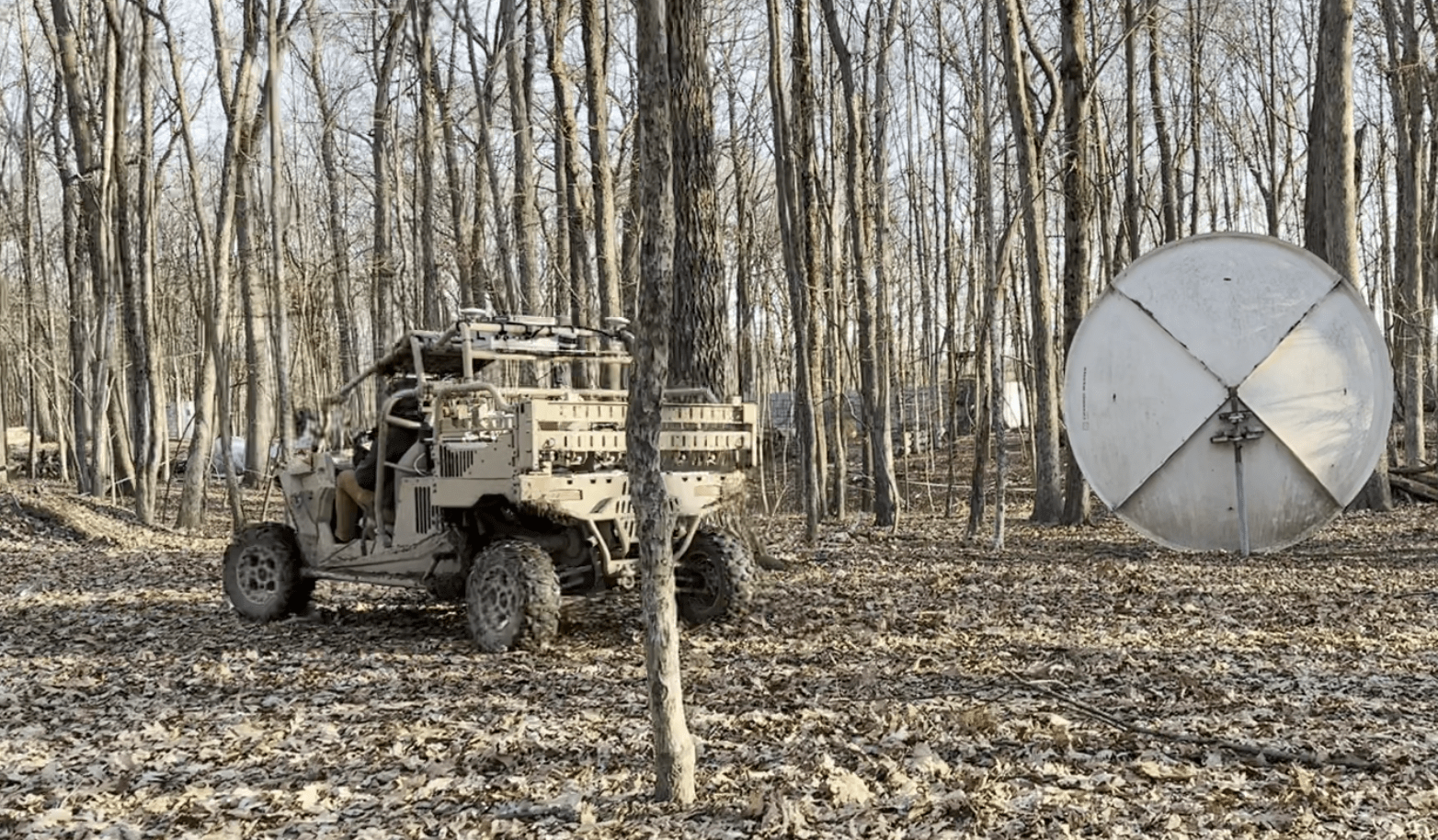}
    \caption{The Polaris MRZR from the Neya Systems. The full-size military vehicle was tested in an off-road forest area. The environment was featureless and the ground was covered in leaves.}
    \label{fig:MRZR}
\end{figure}

\subsection{Off-road Vehicle}
DRIFT also supports full-size vehicles with a single shaft encoder. We evaluate DRIFT on data gathered with the support of Neya Systems from their full-size military test vehicle, the Polaris MRZR. The data was collected with support from Neya Systems at two of their off-road test facilities. As shown in Fig.~\ref{fig:MRZR}, the test facility is a featureless forest area with leafy ground. In this experiment, we obtain GPS signals as approximations to the ground truth positions. 

% The MRZR vehicle contains only a 1-dimensional shaft encoder on the rear transmission shaft. As a result, we follow ~\eqref{eq:pseudo-velocity} to add pseudo-measurements to the lateral and vertical velocity measurements.

Compared to a motion capture system, GPS provides less accurate positional signals. As a result, we compute the final drift and drift percentage in Table~\ref{tab:offroad_quantitative} instead of the RPE. We define the drift percentage as $\text{Final Drift}/\text{Trajectory Length}$. From the table, we see that DRIFT achieves low drift percentages across the three sequences and outperforms the MEKF. 

\begin{figure*}[ht]
    \centering
    \subfloat{
        \includegraphics[width=0.36\textwidth]{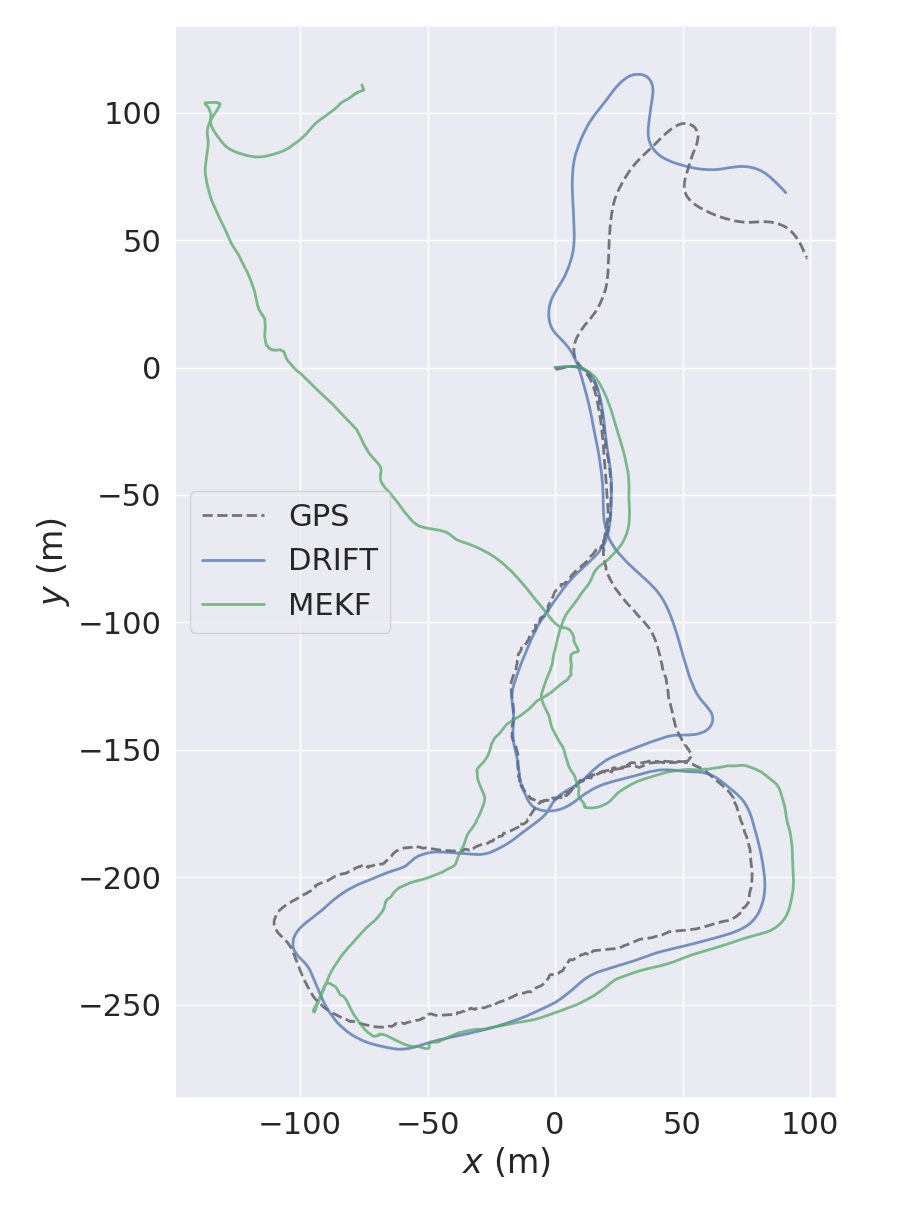}}
    \subfloat{
        \includegraphics[width=0.49\textwidth]{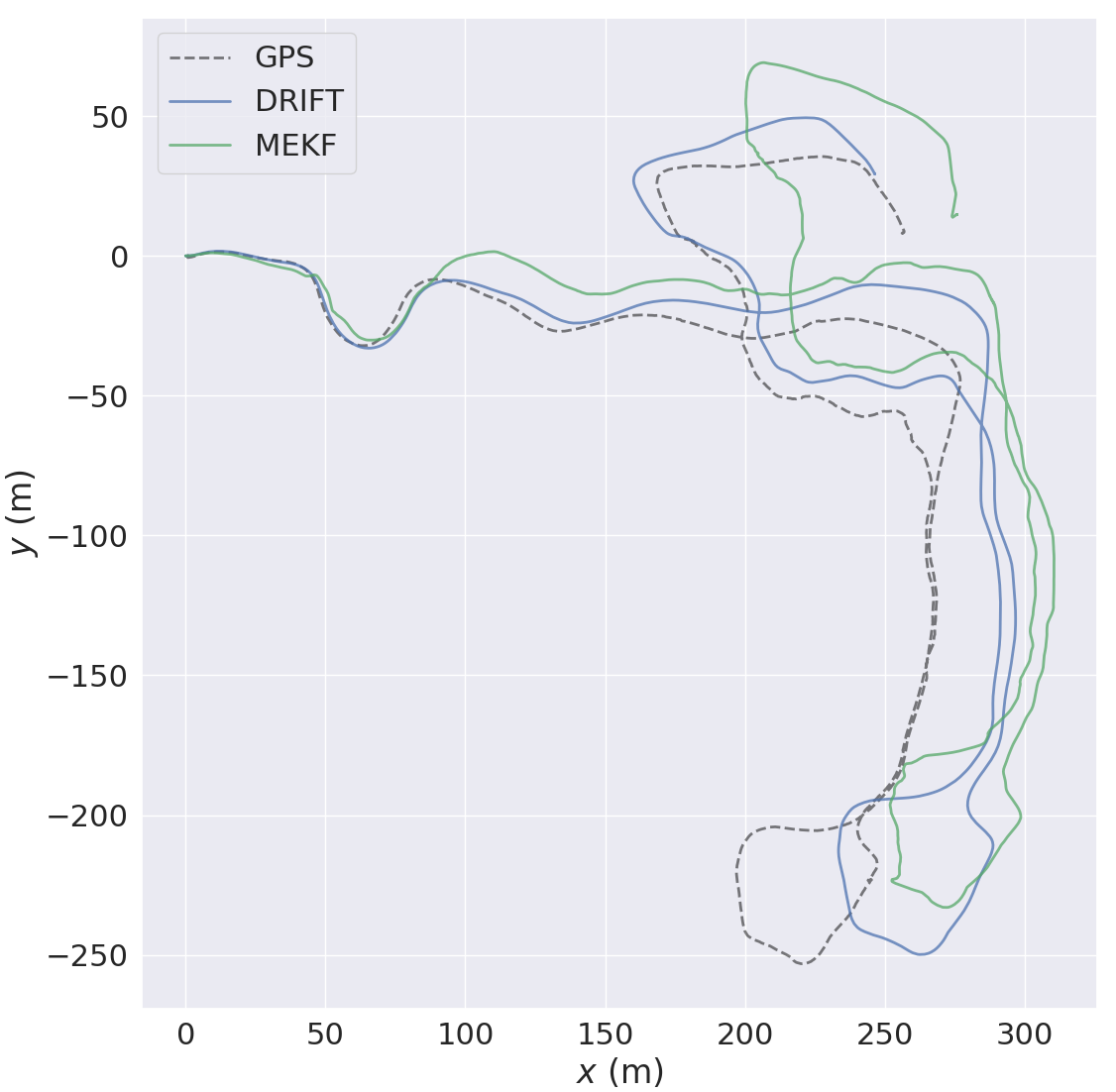}}
        % \label{fig:offraod_leader_follow}}
     \caption{Left: \texttt{1_follow}. Right: \texttt{2_leader_follow}. Bird's-eye views of two sequences from the off-road data set. The data set was collected using a full-size military test vehicle in a featureless forest area. Qualitatively, we see DRIFT outperforms MEKF in both sequences. Since drift only relies on proprioceptive measurements, it can serve as a robust odometry source for modern SLAM systems in such featureless environments.}
     \label{fig:offroad_follow}
\end{figure*}

Fig.~\ref{fig:offroad_follow} 
% and~\ref{fig:offraod_leader_follow} 
shows the bird's-eye view of two off-road sequences, with trajectory lengths of $1.48$ km and $1.26$ km. DRIFT performs significantly better than MEKF in both sequences. This experiment demonstrates that with only an IMU and shaft encoder as inputs, DRIFT can produce highly accurate odometry results for full-size vehicles in off-road environments. These off-road environments are often featureless or with repetitive features. Since DRIFT only relies on proprioceptive measurements, it can serve as a reliable odometry source for perception systems in such scenarios.

% \begin{figure}[ht]
%     \includegraphics[width=0.99\columnwidth]{figures/neya_follow.png}
%     \caption{}
%     \label{fig:offroad_follow}
% \end{figure}

\begin{table*}[t]
\caption{Final drift and drift percentage of the off-road vehicle data set, where the full-size vehicle operates in a forest area for more than 1 kilometer. The drift percentage is computed by $\text{(Final Drift)}/\text{(Trajectory Length)}$.}
\begin{center}
\footnotesize
\begin{tabular}{l l | c c c | c}
    \toprule
     & &  1_follow	& 2_leader_follow & 3_long_run & \textbf{Avg}\\
     \midrule 
      & Trajectory & \multirow{2}{*}{1481.9} & \multirow{2}{*}{1264.0} & \multirow{2}{*}{1785.4} & \multirow{2}{*}{1510.4}\\
      & Length ($\m$) &&&\\
    \midrule
      & Duration ($\sec$) & 317.0 & 424.7 & 605.8 & 449.2\\
    \midrule    
    MEKF~\cite{sola2017quaternion}& Final Drift ($\m$) & 187.4 & 29.6 & 392.1 & 203.0\\
          & Drift Percentage (\%)& 12.7 & 2.3 & 22.0 & 12.3\\
    \midrule
    DRIFT & Final Drift ($\m$) & \textbf{30.2} & \textbf{25.8} & \textbf{97.3} & \textbf{51.1}\\
          & Drift Percentage (\%)& \textbf{2.0} & \textbf{2.0} & \textbf{5.5} & \textbf{3.18}\\

    % \midrule
    % MEKF~\cite{sola2017quaternion} & Trans. ($\m/\m$) & 0.0777 & 0.1363 & 0.1971 & 0.1427 & 0.0881 & 0.1513 & 0.1255 & 0.1245 & 0.1062 & 0.1277 & 0.0384\\
    %    (Same param)   & Rot. ($\degree/\m$) & 1.5364 & 4.7861 & 3.8169 & 1.9029 & 2.2957 & 2.2517 & 2.0711 & 2.6845 & 1.8538 & 2.5777 & 1.0650\\
    \bottomrule
\end{tabular}
\label{tab:offroad_quantitative}
\end{center}
\vspace{-5mm}
\end{table*}

\subsection{Marine Robot}
For marine robots, ground-reference velocity can be obtained via a DVL. The DVL is an acoustic sensor that emits directional acoustic beams to the seabed. The seabed-referenced velocity is then computed using the Doppler effect of the reflected acoustic signals. This enables us to correct the predicted state using the velocity correction method described in Sec.~\ref{sec:velocity_correction}.

\begin{figure*}[t]
    \centering
    \includegraphics[width=0.35\textwidth,clip]{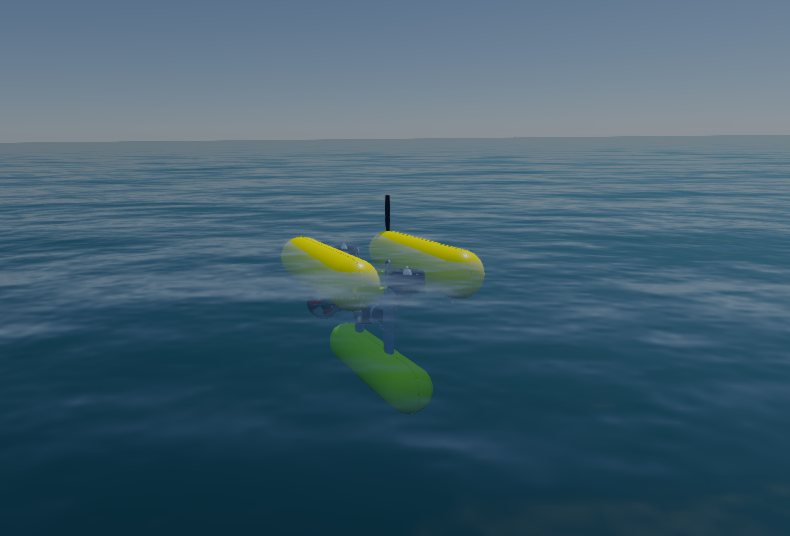}
    \includegraphics[width=0.48\textwidth,trim={0 0.5cm 0 0},clip]{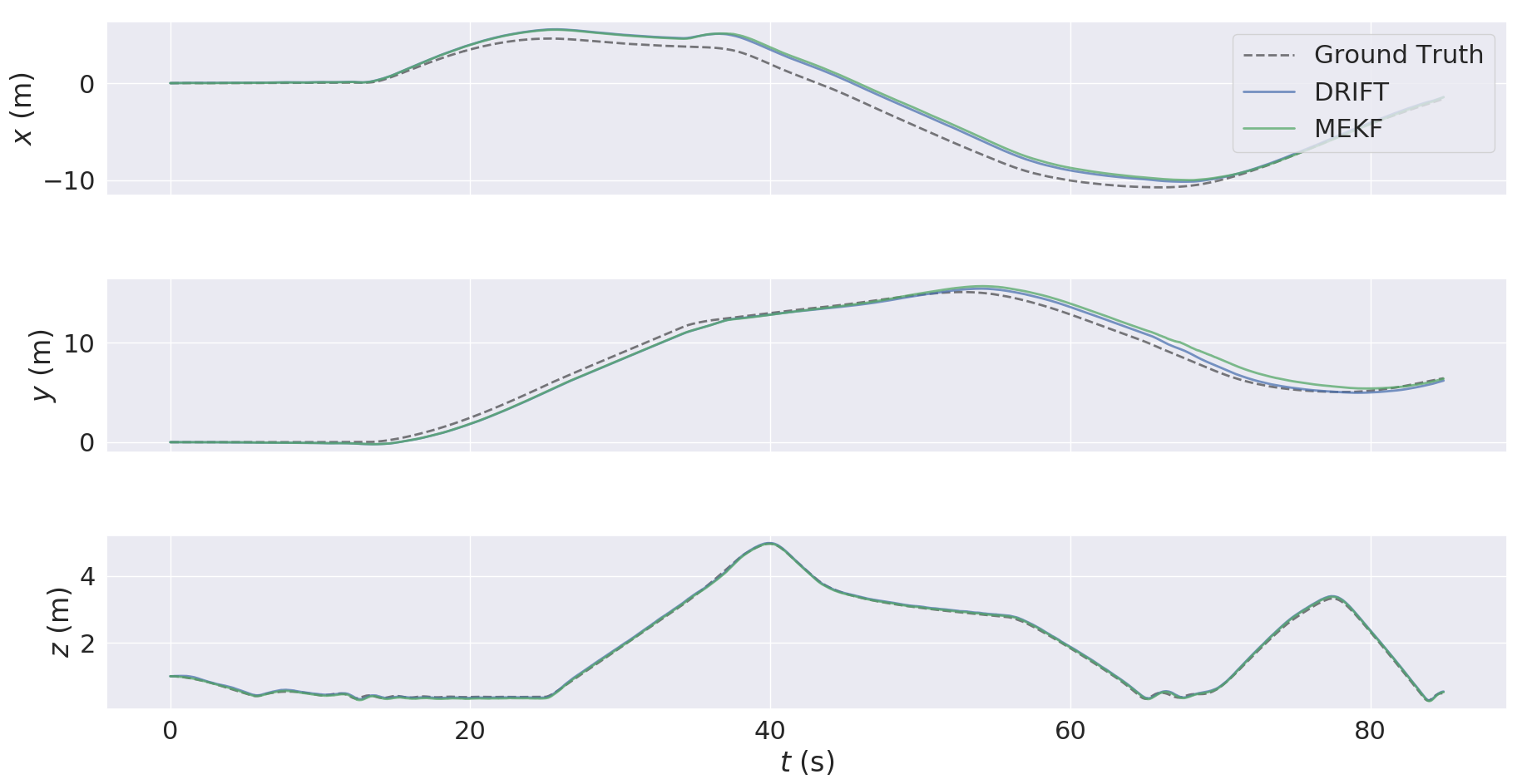}
    \includegraphics[width=0.35\textwidth,clip]{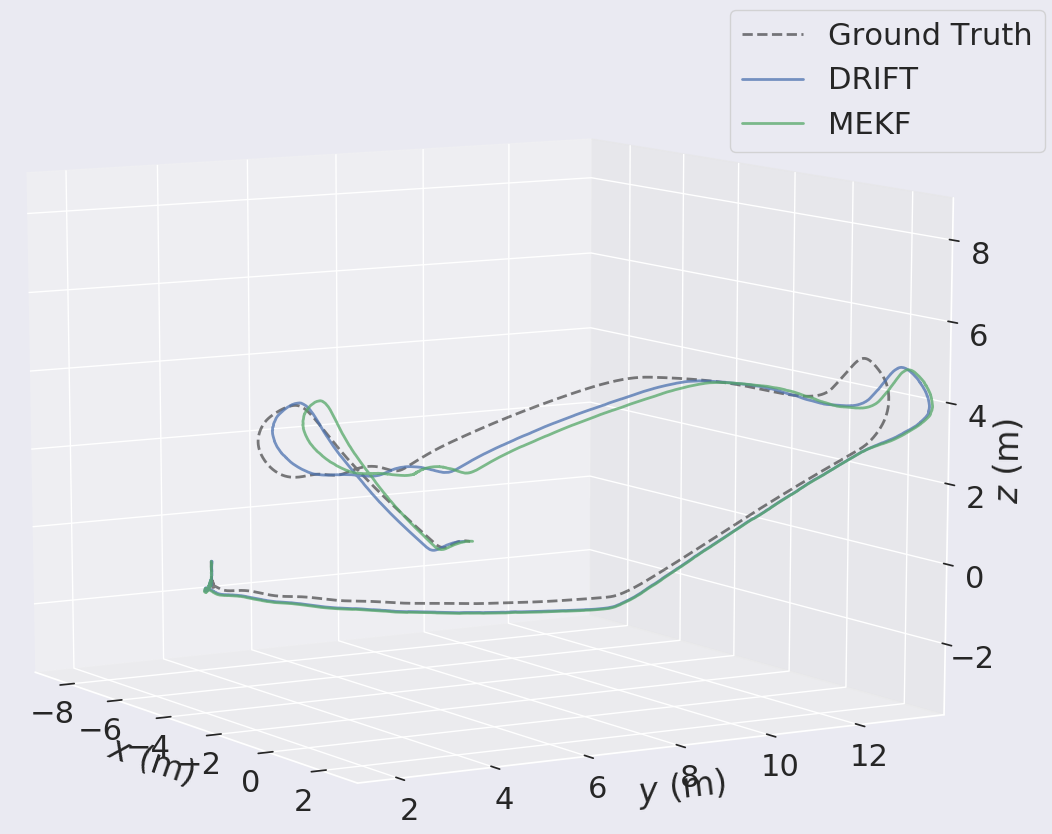}
    \includegraphics[width=0.48\textwidth,trim={0 0 0 0.7cm},clip]{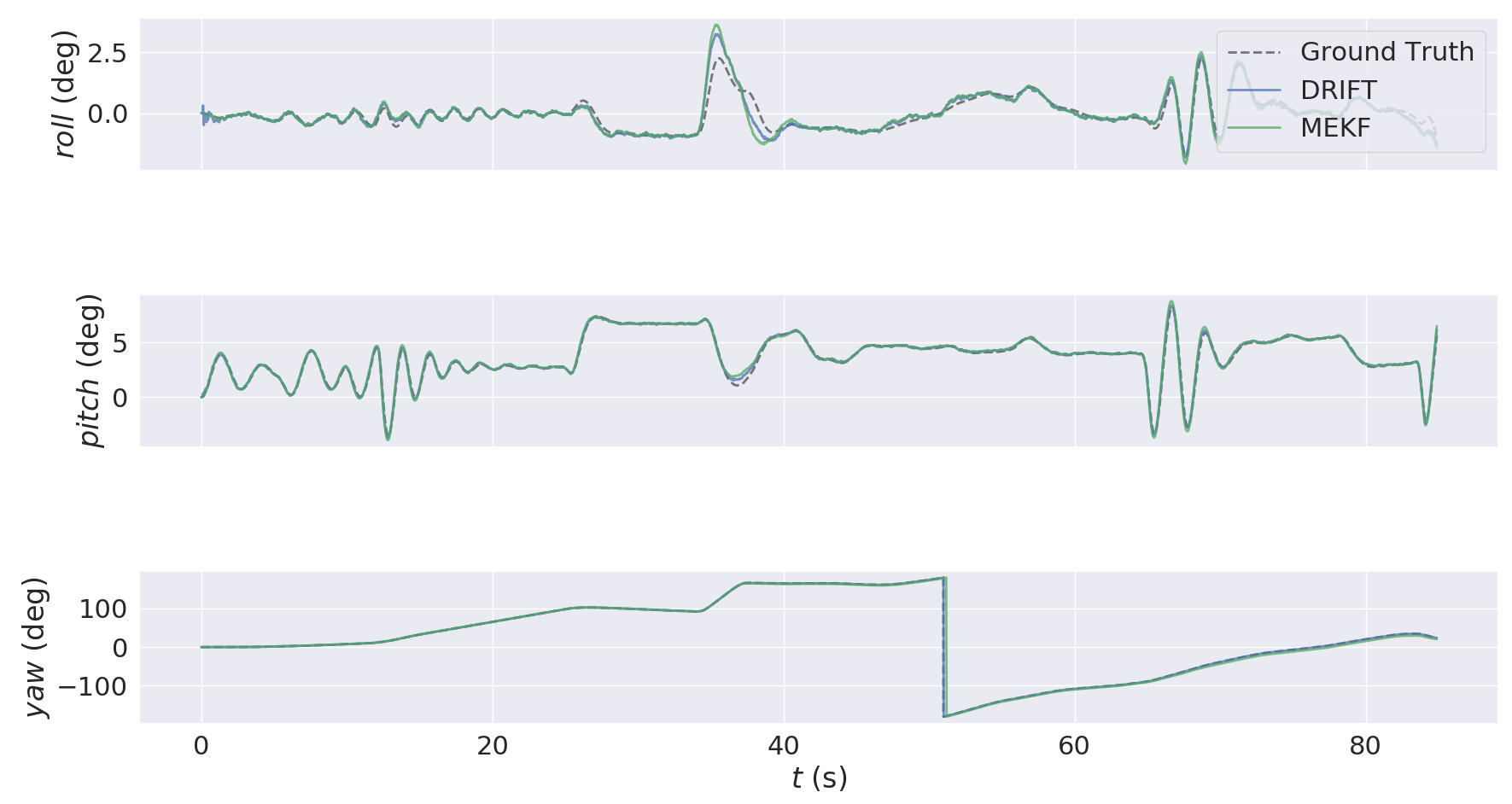}
        
     \caption{Top Left: A Girona500 AUV simulated in the Stonefish simulator~\citep{stonefish}. Bottom Left: The 3D estimated trajectories from the underwater data set. Top Right: The estimated trajectories projected to the $x$, $y$, and $z$ axes. From the plots, we see DRIFT performs slightly better than MEKF. Bottom Right: The estimated roll, pitch, and yaw angles. We observe the roll and pitch estimations converge to the true values after a perturbation around $36 \sec$.}
    %  \squeezeup
     \label{fig:underwater_results}
\end{figure*}

To verify the efficacy of the proposed framework on marine robots, we set up experiments in an open-sourced marine simulation software, Stonefish~\citep{stonefish}. An IQUA Robotics Girona500 Autonomous Underwater Vehicle (AUV) was simulated with an IMU and a DVL attached to the robot, as shown in Fig.~\ref{fig:underwater_results}. The sensor data is corrupted by unbiased Gaussian noises. We used the noise parameters from the VectorNav VN-100 IMU manual to assign the simulation noise parameters. Specifically, we applied $\sigma = \rho\cdot\sqrt{f_s}$ to recover the standard deviation, $\sigma$, from the noise density, $\rho$, and operating frequency, $f_s$. Table~\ref{tab:underwater_sim_noise} lists the noise parameters used in the simulation. We set the IMU and DVL frequencies to $200 \Hz$ and $20 \Hz$.

Fig.~\ref{fig:underwater_results} shows the estimated trajectories from the underwater data set. Qualitatively, we see both DRIFT and MEKF perform well on this sequence. In particular, DRIFT performs slightly better in position estimation in the $x$ and $y$ axes, as well as the roll and pitch estimation. Moreover, the roll and pitch estimations converge to the true values after a perturbation around $36 \sec$, which agrees with our observability analysis in Sec.~\ref{sec:velocity_correction}. 
% This experiment confirms DRIFT's support for marine robotics in an ideal simulation environment. 

% \begin{figure}[ht]
%     \includegraphics[width=0.99\columnwidth]{figures/stonefish.png}
%     \caption{}
%     \label{fig:stonefish}
% \end{figure}

\begin{table}[t]
\caption{Noise parameters used in the underwater simulation.}
\begin{center}
\footnotesize
\begin{tabular}{l | l}
    \toprule
    Measurement Type & Noise Parameter \\
    \midrule
    Angular Velocity & 0.0035 $\degree/\sec/\sqrt{\Hz}$\\
    Linear Acceleration & 0.0014 $\m/\sec/\sqrt{\Hz}$ \\
    DVL Sensor & 0.02626 $\m/\sec$ \\
    \bottomrule
\end{tabular}
\label{tab:underwater_sim_noise}
\end{center}
\end{table}

% \begin{figure*}[ht]
%     \subfloat{
%         \includegraphics[width=0.38\textwidth]{figures/underwater.png}
%         \label{fig:}}
%     \subfloat{
%         \includegraphics[width=0.58\textwidth]{figures/underwater_xyz.png}
%         \label{fig:}}
%      \caption{}
%      \label{fig:underwater_results}
% \end{figure*}

\subsection{Runtime Analysis}
We perform runtime evaluations using a personal laptop with an Intel i5-11400H CPU and an NVIDIA Jetson AGX Xavier (CPU). The Jetson AGX Xavier is commonly used by many robotic platforms. We record the processing time for every propagation and correction method in the InEKF filter and the gyro filter, as presented in Table~\ref{tab:runtime}. DRIFT can operate at an extremely high frequency using CPU-only computation, even on the resourced-constrained Jetson AGX Xavier. For the optional contact estimator, the inference speed on an NVIDIA RTX 3090 GPU is approximately 1100 Hz, and the inference speed on a Jetson AGX Xavier (GPU) is around 830 Hz after TensorRT optimization.

\begin{table}[t]
    \centering
    \caption{Runtime evaluation over datasets on a laptop CPU and an NVIDIA Jetson AGX Xavier. For the Jetson AGX Xavier, all computations are operated on the CPU only.}
    \footnotesize
    \begin{tabular}{l|c|c|c|c}
        & \multicolumn{2}{c}{i5-11400H} & \multicolumn{2}{c}{AGX Xavier (CPU)} \\
        \toprule
         Unit: $\mu s$ &mean &std & mean & std \\
        \toprule
        \textbf{InEKF} \\
        \hline
        propagation & 11.33 & 4.00 & 18.35 & 4.19 \\
        propagation with contact & 10.32 & 4.76 & 22.56 & 7.21\\
        velocity correction & 9.91 & 4.80 & 18.46 & 6.66 \\
        contact correction & 17.46 & 9.78 & 29.39 & 13.07\\
        \hline
        \textbf{Gyro Filter}\\
        \hline
        propagation & 2.57 & 3.46 & 3.96 & 2.28\\
        correction & 2.85 & 2.89 & 4.64 & 4.40 \\
    \end{tabular}
    \label{tab:runtime}
\end{table}

\section{Discussion and future work}
\label{sec:discussion}
One limitation of DRIFT is the assumption of nonholonomic constraints. These constraints can be detached from the robot's actual behavior. Learning such constraints provides a way to use sensory inputs instead of assumptions~\cite{brossard2019rins,brossard2020aiimu,pmlr-v164-lin22b}. Moreover, the nonholonomic constraints are violated when the robot drifts. Slip detection and friction estimation are challenging and necessary tasks for robot state estimation~\cite{yu2022fully}.

Instead of relying on the nonholonomic assumption, marine robots can obtain ground-referenced velocity measurements from a DVL. However, it is worth noticing that a reliable velocity measurement requires a sufficient number of emitted beams to have a line of sights on the seabed, known as \textit{bottom-lock}~\citep{rudolph2012doppler}. If the robot operates at an extreme roll and pitch angle, or if the robot is too far away from the seabed, the bottom-lock might be lost. A typical DVL can obtain bottom-lock for ranges up to 500 m, depending on the frequency of the sensor~\citep{leonard2016autonomous}. In this work, however, we assumed the DVL can always obtain the bottom-lock.

Lastly, DRIFT is a real-time dead-reckoning system. While dead-reckoning might be sufficient for some applications, in many cases it can not replace a localization or SLAM system, where global consistency is the objective. As a result, we position DRIFT as an odometry module of a perception system or a dead-reckoning system for applications that do not require global consistency. A natural next step will then be to incorporate DRIFT into a visual SLAM system for real-time globally consistent state estimation. 

\section{Conclusion}
\label{sec:conclusion}
We developed DRIFT, an open-source, real-time proprioceptive state estimator for multiple robotic platforms. DRIFT is based on the invariant Kalman filtering, leading to significantly better consistency in robot state estimation tasks. Optionally, DRIFT provides two additional modules, a contact estimator and a gyro filter, to assist operations on low-cost robots. We evaluate DRIFT using various robotic platforms with real-world experiments, including a legged robot, an indoor service robot, a field robot, and a full-size vehicle. Additionally, we report simulation results of an underwater vehicle. These experiments verify DRIFT's capability of being a reliable odometry source, even for long-horizon operations. Future directions include extending the framework to support a more general equivariant filter~\cite{van2020equivariant,fornasier2022equivariant}, broader incorporation with symmetry-preserving perception and control algorithms~\cite{ghaffari2022progress}, and incorporating DRIFT into a more comprehensive localization system for global consistency.

{\footnotesize
% \balance
\bibliographystyle{bib/IEEEtranN}
\bibliography{bib/strings-abrv,bib/ieee-abrv,bib/refs}
}

\begin{IEEEbiography}[{\includegraphics[width=1in,height=1.25in,clip,trim={3.1cm 0 3.1cm 0.0cm},keepaspectratio]{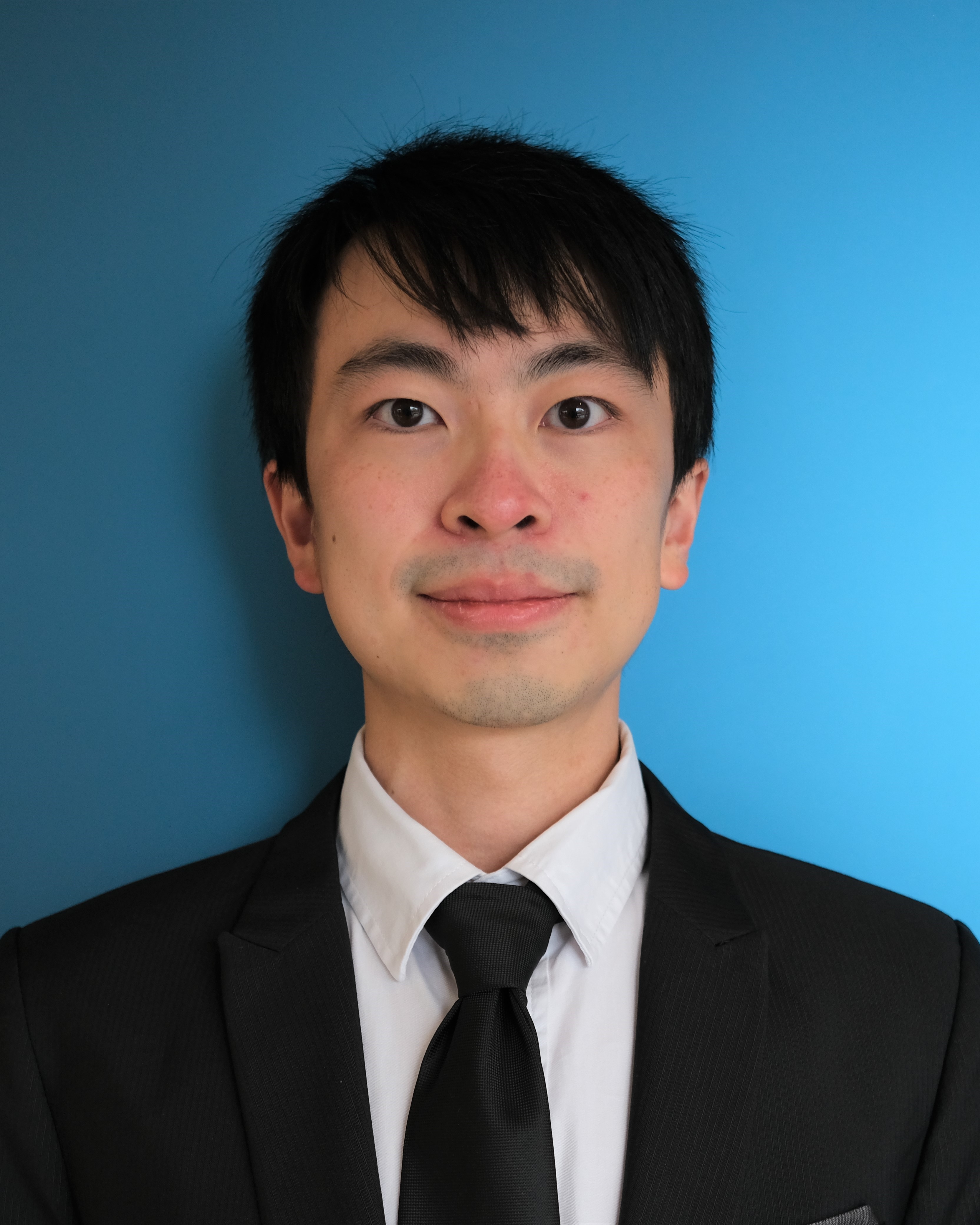}}]{Tzu-Yuan (Justin) Lin} received the B.S. degree in Mechanical Engineering from National Taiwan University, Taipei, Taiwan, in 2017 and the M.S. degree in Robotics from the University of Michigan, Ann Arbor, MI, USA, in 2020. He is currently a Ph.D. candidate in Robotics under the supervision of Professor Maani Ghaffari at the University of Michigan, Ann Arbor.
His research interests include robot perception, simultaneous localization and mapping (SLAM), and computer vision.
\end{IEEEbiography}

\begin{IEEEbiography}[{\includegraphics[width=1in,height=1.25in,clip,trim={3.1cm 0 3.1cm 3.5cm},keepaspectratio]{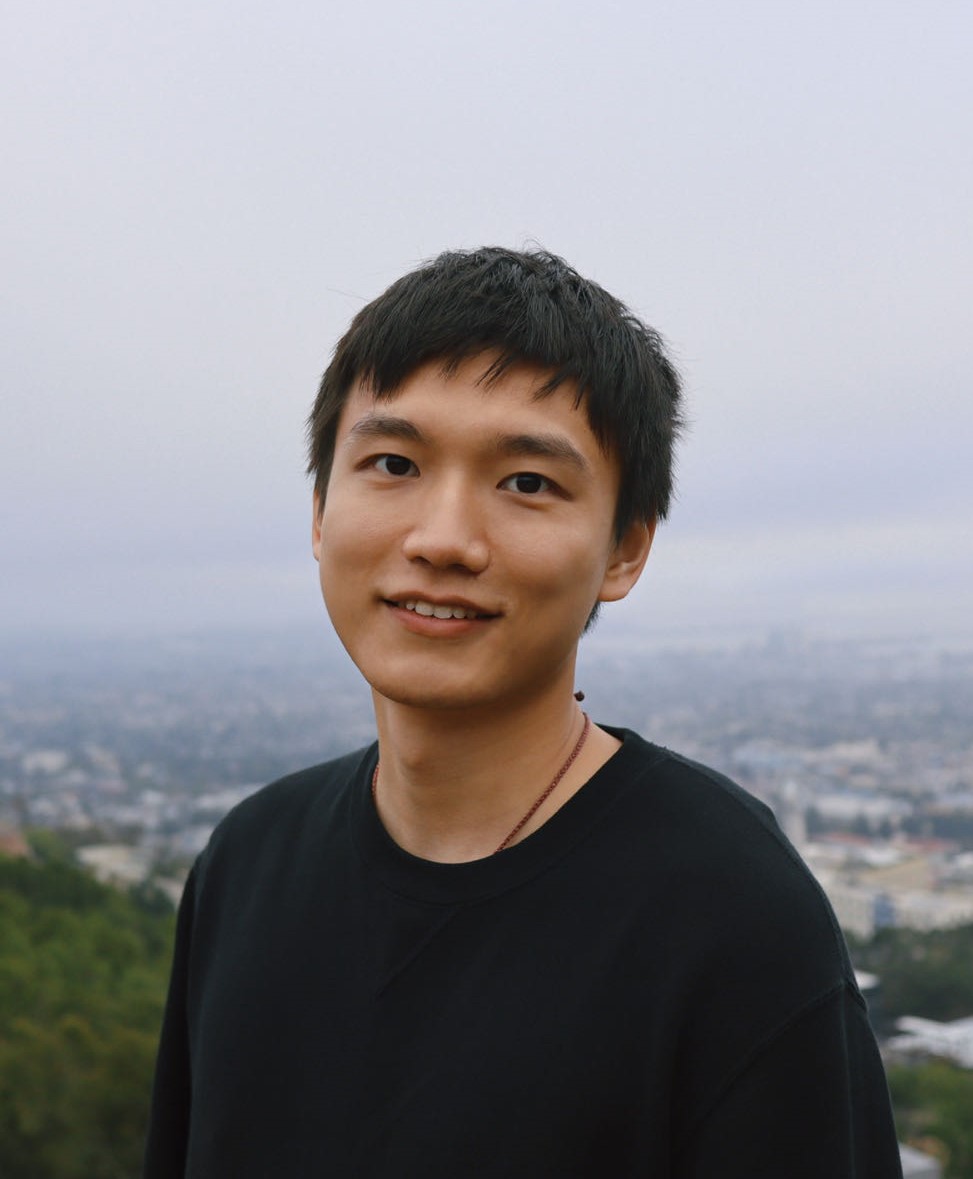}}]{Tingjun Li}
received the Master's degree in Robotics from the University of Michigan, Ann Arbor, MI, USA, in 2022. He was working as a Research Assistant in the CURLY Lab under the supervision of Professor Maani Ghaffari. His interests include simultaneous localization and mapping (SLAM) and the implementation of related algorithms. He is now working at Amazon.com, Inc. as a Software Dev Engineer.
\end{IEEEbiography}

\begin{IEEEbiography}[{\includegraphics[width=1in,height=1.25in,clip,trim={3.1cm 0 3.1cm 3.5cm},keepaspectratio]{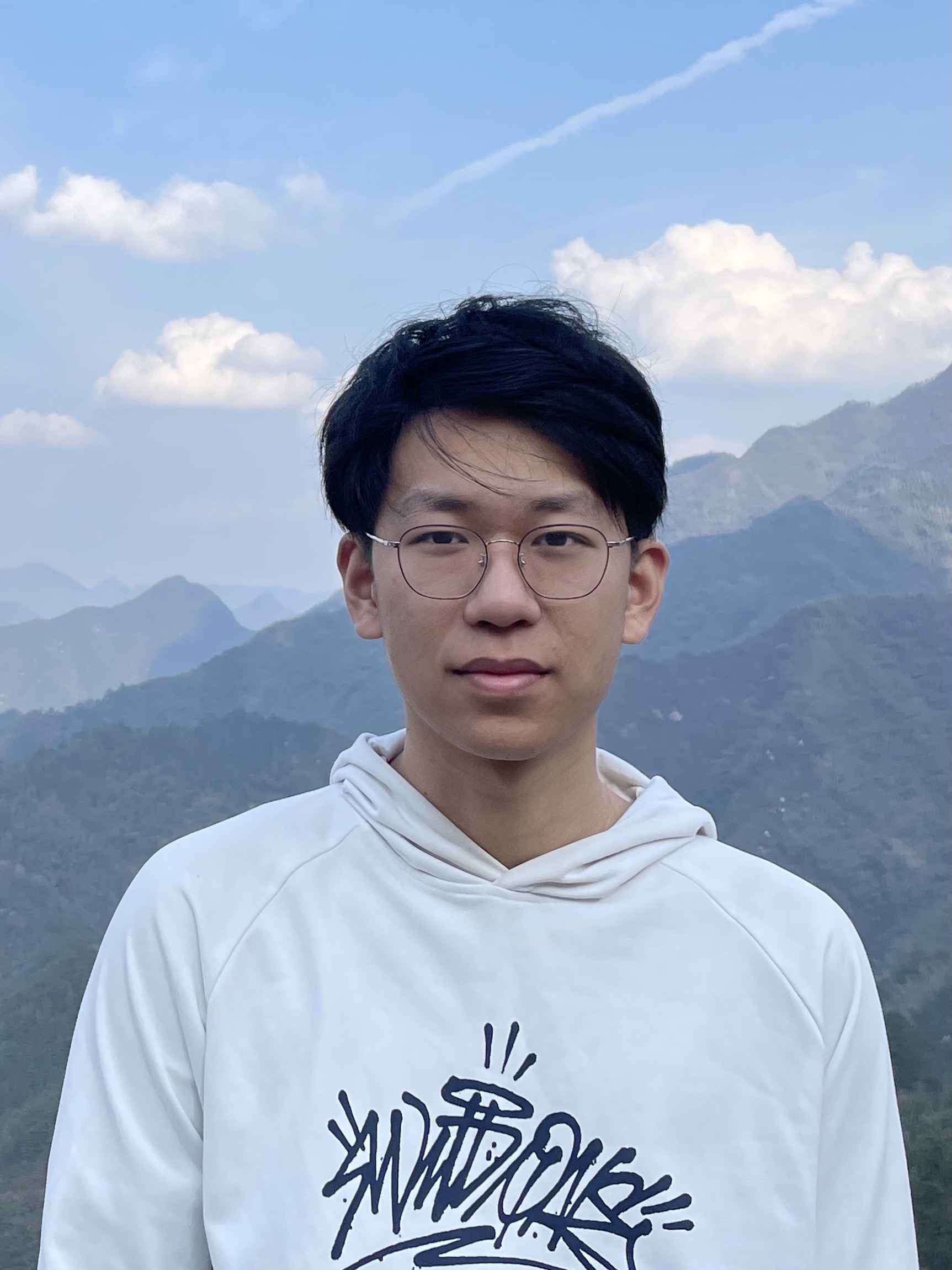}}]{Wenzhe Tong} received the M.S.E. degree in Electrical and Computer Engineering from University of Michigan, Ann Arbor, MI, USA, in 2023. He was working in the CURLY Lab under the supervision of Professor Maani Ghaffari. He is currently a first-year Ph.D. pre-candidate in Robotics under the supervision of Professor Xiaonan Huang at the HDR Lab of the University of Michigan, Ann Arbor. His research interests include robot autonomy, simultaneous localization and mapping (SLAM), and state estimation. 
\end{IEEEbiography}

\begin{IEEEbiography}[{\includegraphics[width=1in,height=1.25in,clip,trim={3.1cm 0 3.1cm 3.5cm},keepaspectratio]{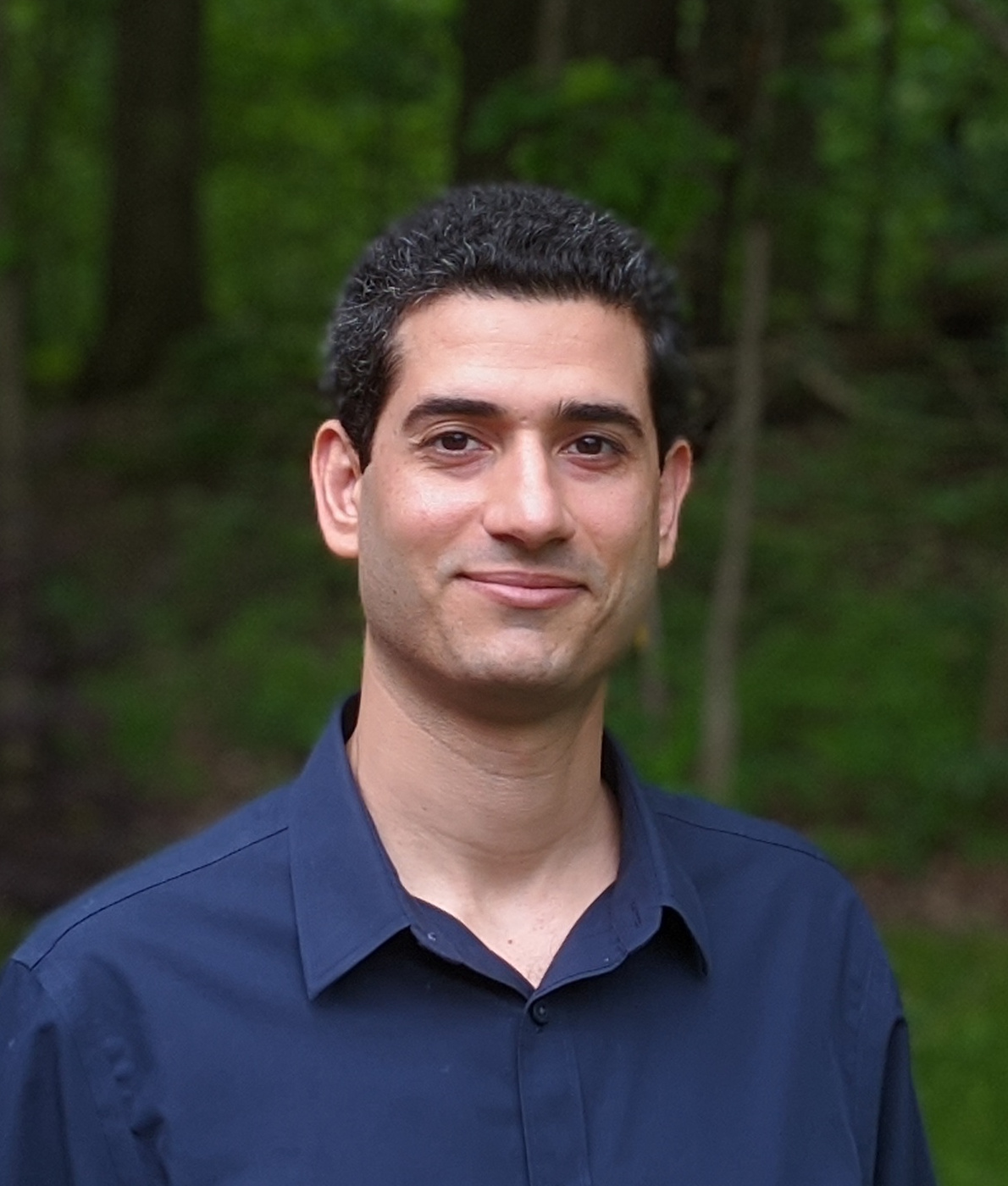}}]{Maani Ghaffari}
received the Ph.D. degree from the Centre for Autonomous Systems (CAS), University of Technology Sydney, NSW, Australia, in 2017. He is currently an Assistant Professor at the Department of Naval Architecture and Marine Engineering and the Department of Robotics, University of Michigan, Ann Arbor, MI, USA, where he directs the Computational Autonomy and Robotics Laboratory (CURLY). His work on sparse, globally optimal kinodynamic motion planning on Lie groups received the best paper award finalist title at the 2023 Robotics: Science and Systems conference. He is the recipient of the 2021 Amazon Research Awards. His research interests lie in the theory and applications of robotics and autonomous systems.
\end{IEEEbiography}

\vfill

\end{document}